\documentclass[preprint,5p,times,twocolumn]{elsarticle}
\usepackage{natbib}
\usepackage{multirow}
\usepackage{color}
\usepackage{multicol}
\usepackage[table]{xcolor}

\usepackage{amssymb}
\usepackage{gensymb}
\usepackage{fontenc}
\usepackage{amsmath}
\usepackage{hyperref}

\usepackage{lineno}

\newcommand{\argmaxA}{\mathop{\mathrm{arg\,max}}}

\begin{document}

\begin{frontmatter}

\title{3D CNN-based classification using sMRI and MD-DTI images for Alzheimer disease studies}
	
\author[msu,labri]{Alexander Khvostikov\corref{cor1}}
\ead{xubiker@gmail.com}

\author[labri,maroc]{Karim Aderghal}
\ead{aderghal.karim@gmail.com}

\author[labri]{Jenny Benois-Pineau}
\ead{jenny.benois-pineau@u-bordeaux.fr}

\author[msu]{Andrey Krylov}
\ead{kryl@cs.msu.ru}

\author[bordeaux2]{Gwenaelle Catheline}
\ead{gwenaelle.catheline@u-bordeaux2.fr}

\author{for the Alzheimer's Disease Neuroimaging Initiative\tnoteref{adni}}

\cortext[cor1]{Corresponding author}

\tnotetext[adni]{Data used in preparation of this article were obtained from the Alzheimer's Disease Neuroimaging Initiative (ADNI) database (\url{http://adni.loni.usc.edu}). As such, the investigators within the ADNI contributed to the design and implementation of ADNI and/or provided data but did not participate in analysis or writing of this report. A complete listing of ADNI investigators can be found at: \url{http://adni.loni.usc.edu/wp-content/uploads/how_to_apply/ADNI_Acknowledgement_List.pdf}.}

\address[msu]{Lomonosov Moscow State University, Department of Computational Mathematics and Cybernetics, Moscow, Russia}

\address[labri]{LaBRI UMR 5800, University of Bordeaux, Bordeaux, France}

\address[maroc]{LabSIV, Université Ibn Zhor, Agadir, Maroc}

\address[bordeaux2]{University Victor Segalen Bordeaux 2, Bordeaux, France}

\begin{abstract}
Computer-aided early diagnosis of Alzheimer’s Disease (AD) and its prodromal form, Mild Cognitive Impairment (MCI), has been the subject of extensive research in recent years. Some recent studies have shown promising results in the AD and MCI determination using structural and functional Magnetic Resonance Imaging (sMRI, fMRI), Positron Emission Tomography (PET) and Diffusion Tensor Imaging (DTI) modalities. Furthermore, fusion of imaging modalities in a supervised machine learning framework has shown promising direction of research. 

In this paper we first review major trends in automatic classification methods such as feature extraction based methods as well as deep learning approaches in medical image analysis applied to the field of Alzheimer's Disease diagnostics. Then we propose our own algorithm for Alzheimer's Disease diagnostics based on a convolutional neural network and sMRI and DTI modalities fusion on hippocampal ROI using data from  the  Alzheimer’s  Disease Neuroimaging  Initiative  (ADNI)  database (\url{http://adni.loni.usc.edu}). Comparison  with a single modality approach shows promising results. We also propose our own method of data augmentation for balancing classes of different size and analyze the impact of the ROI size on the classification results as well.
	
\end{abstract}


\end{frontmatter}


\section{Introduction}
\label{S:Introduction}

Alzheimer’s Disease (AD) is the most common type of dementia. It is characterized by degeneration of brain cells which results in changes of brain structures noticeable on images form different imaging modalities e.g. sMRI, DTI, PET. With the development of machine learning approaches, research on computer-aided diagnostics (CAD) has become very much intensive \cite{volumetric:VBM},\cite{c2.2_3D_CNN_USA, c4_sparse_encoder_Korea, c8_medical_UK, c5_demnet_Philippines}. 

Images of different modalities such as structural and functional magnetic resonance imaging (sMRI, fMRI), positron emission tomography (PET) and diffusion tensor imaging (DTI) scans  can be used for early detection of the disease.

The majority of earlier works were focused on the volumetric-based approaches that perform comparison of anatomical brain structures assuming one-to-one correspondence between subjects. The wide-spread voxel-based morphometry (VBM) \cite{volumetric:VBM} is an automatic volumetric method for studying the differences in local concentrations of white and gray matter and comparison of brain structures of the subjects to test with reference normal control (NC) brains. Tensor-based morphometry (TBM) \cite{volumetric:TBM} was proposed to identify local structural changes from the gradients of deformations fields when matching tested brain and the reference healthy NC. Object-based morphometry (OBM) \cite{volumetric:OBM} was introduced for shape analysis of anatomical structures.

In general, the automatic classification on brain images of different modalities can be applied to the whole brain \cite{c2.2_3D_CNN_USA, c4_sparse_encoder_Korea, c8_medical_UK, c5_demnet_Philippines},
or performed  using the domain knowledge on specific regions of interest (ROIs). Structural changes in some  structures e.g. hippocampal ROI are strongly correlated to the disease \cite{pierrick}. The changes in such regions are considered as AD biomarkers.  

Advances in computer vision and content-based image retrieval research made penetrate the so-called feature-based methods into classification approaches for (AD) detection\cite{f1_Jenny_MKL,f3_Jenny_pcc,f5_Jenny}. 
The reason for this is in inter-subject variability, which is difficult to handle in VBM. On the contrary,  the quantity of local features which can be extracted form the brain scans together with captured particularities of the image signal allowed an efficient classification with lower computational workload \cite{f5_Jenny}. The obtained feature vectors are classified using machine learning algorithms.

Lately with the development of neural networks the feature-based approach became less popular and is gradually replaced with convolutional neural networks of different architectures.

In the present paper we give a substantial overview of recent trends in classification of different brain imaging modalities in the problem of computer-aided diagnostics of Alzheimer disease and its prodromal stage, i.e. mild cognition impairment (MCI) and propose our own algorithm for this purpose. The algorithm is based on the recent trend in supervised machine learning such as Deep Convolutional Neural Networks (CNNs). We propose an adapted architecture of a CNN for classification of 3D volumes of hippocampal ROIs and explore fusion of two modalities sMRI and DTI available for the same cohort of patients. In our work we use a subset of ADNI database (\url{http://adni.loni.usc.edu}).

The paper is organized as follows. In Section \ref{sec:Classification Approaches} we overview the recent trends in classification of brain images in the problem of AD detection. Main feature-based approaches are presented in section  \ref{sec:texture_based}. In section \ref{sec:networks} we compare different approaches based on neural networks. Particular attention is paid in each case to fusion of modalities. All reviewed approaches are compared in Table \ref{comparison_table}. In section \ref{Our method} we present the proposed method of classification with 3D CNNs. In section \ref{sec:Results} results of the method are presented. Section \ref{S:Conclusion} contains discussion and conclusion of our work and outlines research perspectives. 
\color{black}

\section{Review of the existing classification methods in the problem of AD detection}
As an alternative to heavy volumetric methods, feature-based approaches were applied in the problem of AD detection using domain knowledge both on the ROI biomarkers and on the nature of the signal in sMRI and DTI modalities which is blurry and cannot be sufficiently well described by conventional differential descriptors such as SIFT\cite{SIFT} and SURF\cite{SURF}. 
\label{sec:Classification Approaches}
\subsection{Feature-based classification}
\label{sec:texture_based}

Feature-based classification can be performed on images of different modalities. Here we compare and discuss the usage of sMRI, DTI and sMRI fusion with other modalities.

\subsubsection{sMRI}

In previous joint work \cite{f3_Jenny_pcc}, Ahmed et al. computed local features on  sMRI scans in hippocampus and posterior cingulate cortex (PCC) structures of the brain. The originality of the work consisted in the usage of Gauss-Laguerre Harmonic Functions (GL-CHFs) instead of traditional SIFT\cite{SIFT} and SURF\cite{SURF} descriptors. CHFs perform image decomposition on the orthonormal functional basis, which allows capturing local directions of the image signal and intermediate frequencies. It is  similar to Fourier decomposition, but is more appropriate in case of smooth contrasts of MRI modality. For each projection of each ROI a signature vector was calculated using a bag-of-visual-words model (BoVWM) with a low-dimensional dictionary with  300 clusters. This led to the total signature length of 1800 per image. Principal component analysis was then applied to reduce the signature length to 278. The signatures then were classified using SVM with RBF kernel and 10-fold cross-validation and reached the accuracy level of 0.838, 0.695, 0.621 for AD/NC, NC/MCI and AD/MCI binary classification problems accordingly on the subset of ADNI database.

\subsubsection{DTI}

This modality is probably the most recent to be used for AD classification tasks. Both  Mean Diffusivity (MD) and Fractional Anisotropy (FA) maps are being explored for this purpose.  
In \cite{g1_graph_ensemble_2017} the authors acquired DTI images of 15 AD patients, 15 MCI patients, and 15 healthy volunteers (NC). After the preprocessing steps the FA map, which is an  indicator of brain connectivity,  was calculated. The authors considered 41 Brodmann areas, calculated the connectivity matrices for this areas and generated a connectivity graph with corresponding 41 nodes. Two nodes corresponding to Brodmann areas are marked with an edge if there is at least one fiber connecting them. Then the graph is described with the vector of features, calculated for each node and characterizing the connectivity of the node neighborhood. Totally each patient is characterized by 451 feature. The vectors were reduced to the size of 430 and 110 using ANOVA-based feature selection approach. All vectors were classified with the ensemble of classifiers (Logistic regression, Random Forest, Gaussian native Bayes, 1-nearest neighbor, SVM) using 5-fold cross-validation. The authors have achieved the 0.8, 0.833, 0.7 accuracy levels for AD/NC, AD/MCI and MCI/NC accordingly on their custom database.

Another methodology is described in \cite{g3_svm_DTI_Korea}. The authors use the fractional anisotropy (FA) and mode of anisotropy (MO) values of DTI scans of 50 patients from the LONI Image Data Archive (\url{https://ida.loni.usc.edu}). After non-linear registration to the standard FA map, the authors calculate the skeleton of the mean FA image as well as MO and perform the second step of registration. After that a Relief feature algorithm is performed on all voxels of the image, relevant ones are used for 10-cross validation training the SVM classifier with RBF kernel. The declared accuracy is 0.986 and 0.977 for classification  AD/MCI, AD/NC accordingly.

\subsubsection{Data fusion}

In \cite{g2_sift_China} authors use a fusion of sMRI and PET images together with canonical correlation analysis (CCA). After preprocessing and aligning images of 2 modalities given the covariance data of sMRI and PET image they find the projection matrices by maximizing the correlation between projected features. Here
\[
X_1 \in R^{d \times n}, X_2 \in R^{d \times n}
\]
are the \(d\)-dimentional sMRI and PET features of \(n\) samples,
\[
\Sigma = \begin{bmatrix}
    \Sigma_{11} &  \Sigma_{12} \\
    \Sigma_{21} &  \Sigma_{22} \\
  \end{bmatrix}
\]
is a covariance matrix, 
\[
(B_1, B_2) = \argmaxA_{(B_1, B_2)} \frac{B_1^T \Sigma_{12} B_2}{\sqrt[]{B_1^T \Sigma_{11} B_1}\sqrt[]{B_2^T \Sigma_{22} B_2}}
\]
are the projection matrices and
\[
Z_1 = B_1^T X_1, Z_2 = B_2^T X_2
\]
are the resulting projections.
The authors construct the united data representation for each patient:
\[
F = [X_1; X_2; Z_1; Z_2] \in R^{4d \times n}
\]
and calculate SIFT descriptors. This descriptors are used to form the BoVW model, the classification is performed using SVM. The achieved accuracy is 0.969, 0.866 and for classifying AD/NC and MCI/NC accordingly on a subset of ADNI database.

Ahmed et al. in \cite{f5_Jenny} demonstrated the efficiency of using the amount of cerebrospinal fluid (CSF) in the hippocampal area calculated by an adaptive Otsu's  thresholding method as an additional feature for AD diagnostics. In \cite{f1_Jenny_MKL} they further improved the result of \cite{f3_Jenny_pcc} by combining visual features derived from sMRI and DTI MD maps with a multiple kernel learning scheme (MKL). Similar to \cite{f3_Jenny_pcc} they selected hippocampus ROIs on the axial, saggital and coronal projections and described them using Gauss-Laguerre Harmonic Functions (GL-CHFs). These features are clustered into 250 and 150 clusters for sMRI and MD DTI modalities and encoded using the BoVW model. Thus they got three sets of features: BoVW histogram for sMRI, BoVW histogram for MD DTI and CSF features. The obtained vectors are classified using MKL approach based on SVM. The achieved accuracy is 0.902, 0.794, 0.766 for AD/NC, MCI/NC and AD/MCI classification on a subset of ADNI database.

\subsection{Classification with neural networks}
\label{sec:networks}

Deep neural networks (DNN) and specifically convolutional NN (CNNs) have become popular now due to their good generalization capacity and available GPU Hardware needed for parameter optimization.  Their main drawback for AD classification  is the small amount of available training  data  and also a low resolution of input images when the ROIs are considered. This problem can be eliminated in several ways: i) by using shallow networks with relatively small number of neurons, ii) applying transfer learning from an  existing trained network or iii) pretraining  some of the layers of the network.

Forming shallow networks kills the idea of deep learning to recognize structures at different scales and reduces the generalization ability of the network, so this methodology has not often been used since recently, despite it has shown decent results \cite{aderghal2017classification}. In this case the classification performance could be enhanced by selecting several ROIs in each image and applying the voting rule. In particular in \cite{ex_2} authors used 7 ROIs in each sMRI image.

One way to enlarge the dataset is to use domain-dependent data augmentation. In case of medical images this often comes down to mirror flipping, small-magnitude translations and weak Gaussian blurring \cite{aderghal2017classification}.

Another way is to use more input data e.g. consider several ROIs instead of one. So Liu et al. in \cite{ex_6} first identify discriminative 50 anatomical landmarks from MR images in a data-driven manner, and then extract multiple image patches around these detected landmarks. After that they use a deep multi-task multi-channel convolutional neural network for disease classification. The authors addressed the problem of classification of patients into NC, stable MCI (sMCI) an progressive MCI (pMCI). The authors used MRI images from ADNI database containing in total 1396 images and achieved 0.518 accuracy in four-class (NC/sMCI/pMCI/AD) classification.

A more simplified idea was proposed by Cheng et al. in \cite{ex_7} as they used a number of 3D convolutional neural networks with 4 layers together with late fusion. With the subset of ADNI database of 428 sMRI images the authors achieved an accuracy value of 0.872 for AD/NC classification.

\subsubsection{Autoencoders}

\begin{figure}[h]
\centering\includegraphics[width=0.7\linewidth]{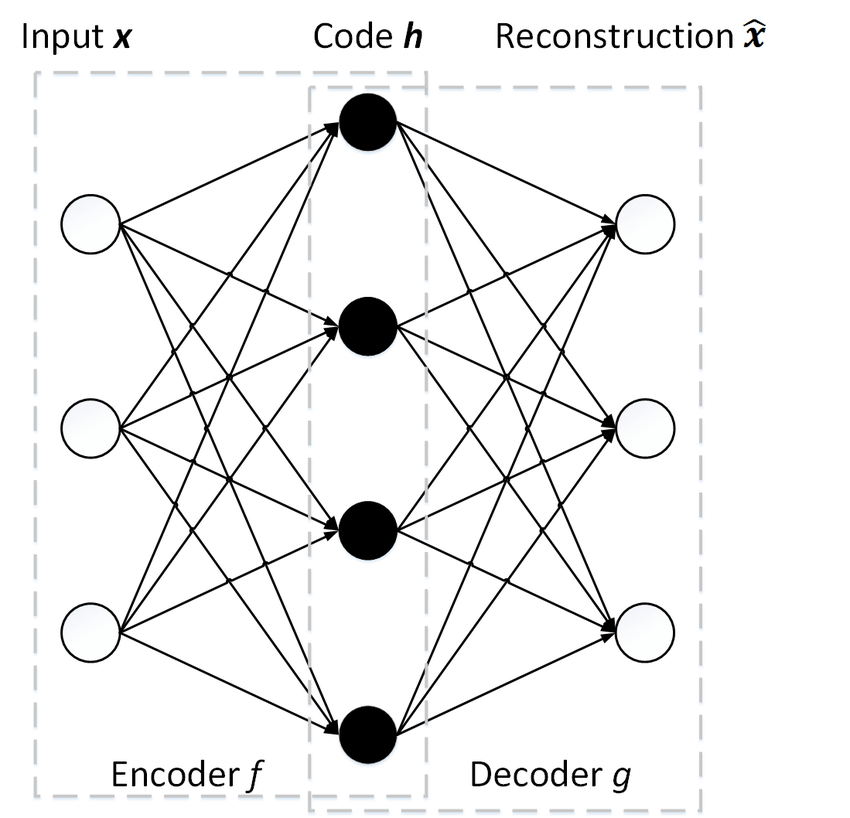}
\caption{Architecture of autoencoder}
\label{fig:autoencoder}
\end{figure}

The idea of pretraining some of the layers in the network is easily implemented with autoencodedrs (AE) or in  image processing tasks more often with convolutional autoencoders (CAE). Autoencoder consists of an input layer, hidden layer and an output layer, where the input and output layers have the same number of units (Fig.\ref{fig:autoencoder}). Given  the input vector \(x \in \mathbb{R}^n \) autoencoder maps it to the hidden representation \(h\):
\[
h = f(Wx+b),
\]
where \(W \in \mathbb{R}^{p \times n} \) are the weights, \(b \in \mathbb{R}^p\) are the biases, \(n\) is the number of input units, \(p\) is the number of hidden units, \(f\) is anon-linear encoder function e.g. sigmoid.
After that the hidden representation \(h\) is mapped back to \(\tilde{x} \in \mathbb{R}^n \):
\[
\widehat{x} = g(\widehat{W}h + \widehat{b}),
\]
where \(\widehat{W} \in \mathbb{R}^{n \times p} \), \(b \in \mathbb{R}^n\), \(g\) is the identity function.
The weights and biases are found by gradient methods to minimize the cost function:
\[
J(W, b) = \frac{1}{N} \sum_{i=1}^{N}{\frac{1}{2} || \widehat{x}^{(i)} - x^{(i)} ||^2},
\]
where\(N\) is the number of inputs.

The overcompleted hidden layer is used to make the autoencoder extracting features.

Introducing spatial constraints with convolutions easily alignes the model of autoencoder to the convolutional autoencoder (CAE) and 3D convolutional autoencoder (3D-CAE).

In \cite{c4_sparse_encoder_Korea} authors added a sparsity constraint to prevent hidden layers of autoencoder from learning the identity function. They use 3D convolutions on the both sMRI and PET modalities and train the autoencoder on random \(5 \times 5 \times 5\) image patches. Maxpooling, fully-connected and softmax layers were applied after autoencoding. Mixing data of sMRI and PET modalities is performed at FC layer. The use of autoencoders allowed the authors using a subset of ADNI database to increase the classification accuracy by 4-6\% and leads to the level of 0.91 for AD/NC classification.

\begin{figure}[h]
\centering\includegraphics[width=1.0\linewidth]{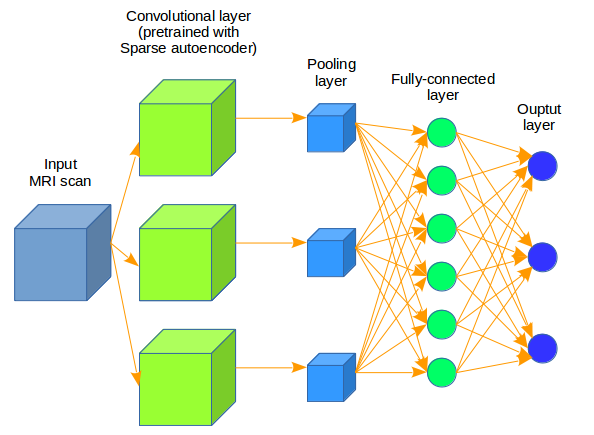}
\caption{Typical CNN architecture with CAE pretraining.}
\label{fig:cnn_1}
\end{figure}

Nearly the same approach with a sparse 3D autoencoder was used in \cite{c8_medical_UK} to classify sMRI images into 3 categories (AD/MCI/NC). The proposed network architecture is shown in Fig.\ref{fig:cnn_1}.  Larger obtained dataset selected from ADNI database and more accurate network parameters configuration allowed the authors to reach the accuracy of 0.954, 0.868 and 0.921 in AD/NC, AD/MCI and NC/MCI determination accordingly.

The authors of \cite{c2.2_3D_CNN_USA} extended the idea of applying autoencoders.  They proposed using three stacked 3D convolutional autoencoders instead of only one. Two fully-connected layers before the softmax were used for a progressive dimension reduction. The usage of stacked 3D CAE allowed the authors to achieve one of the best accuracy levels on 2265 images from ADNI database: 0.993, 1, 0.942 for AD/NC, AD/MCI and MCI/NC classification using sMRI images only. 

\subsubsection{Transfer Learning}

\begin{figure}[h]
\centering\includegraphics[width=1.0\linewidth]{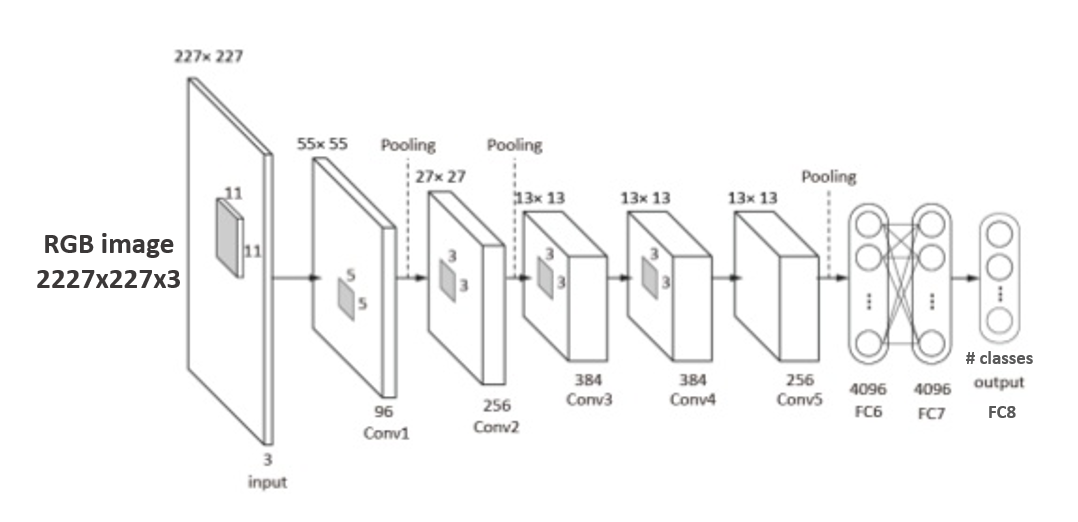}
\caption{AlexNet architecture. Includes 5 convolutional layers and 3 fully-connected layers.}
\label{fig:alexnet}
\end{figure}

Transfer learning is considered as the transfer of knowledge from one learned task to a new task in machine learning. In the context of neural networks, it is transferring learned features of a pretrained network to a new problem. Glozman and Liba in \cite{c1_report_Standford} used the widely known AlexNet \cite{alexnet}, pretrained on the ImageNet benchmark and fine-tuned the last 3 fully-connected layers (Fig.\ref{fig:alexnet}). The main problem of transfer learning is the necessity to transform the available data so that it corresponds to the network input. In \cite{c1_report_Standford} the authors created several 3-channel 2D images from the 3D input of sMRI and PET images by choosing central and nearby slices from axial, coronal and saggital projections. They  then interpolated the slices to the size \(227 \times 227\) compatible with AlexNet. Naturally one network was used for each projection. To augment the source data only mirror flipping was applied. This transfer learning based approach allowed the authors to reach 0.665 and 0.488 accuracy on 2-way (AD/NC) and 3-way (AD/MCI/NC) classifications accordingly on a subset of ADNI database.

In \cite{ex_3} authors apart from using the transfer learning technique proposed a convolutional neural network by involving Tucker tensor decomposition for classification of MCI subjects. The achieved accuracy on a subset of ADNI database containing 629 subjects is 0.906.

\subsubsection{2D convolutional neural networks}

In \cite{c7.1_DeepAd_Canada}, \cite{c7.3_DeeapAd_Canada}, \cite{c7.2_DeeapAd_Canada} the authors compared the classification of structural and functional MRI images using one of the lightest  Deep architectures, the LeNet-5 architecture. They  transformed  the source 3D and 4D (in case of fMRI) data to a batch of 2D images. LeNet-5 consists of two convolutional and two fully-connected layers. The reached level of accuracy for 2-class classification (AD/NC) was 0.988 for sMRI and 0.999 for fMRI images.

Billones et al. proposed in \cite{c5_demnet_Philippines} to use a modified 16-layered VGG network \cite{VGG} to classify sMRI images. The key feature of this paper was to use 2D convolutional network to classify each slice of source data separately. The authors selected 20 central slices for each image and the final score was calculated as the output of the last softmax layer of the network. The accuracy of each slice among all images was also studied, 17 slices were selected as representative, 3 slices (the first and two last slices in the image sequence) demonstrated lower level of accuracy. All in all authors reached a very good accuracy level: 0.983, 0.939, 0.917 for AD/NC, AD/MCI and MCI/NC classification using 900 sMRI images from a subset of ADNI database.

In \cite{Aderghal} Aderghal et al. used 3 central slices in each projection of a hippocampal ROI. The network architecture represented three 2D convolutional networks (one network per projection) that were joined in the last fully-connected layer. The reached accuracy for AD/NC, AD/MCI and MCI/NC classification is 0.914, 0.695 and 0.656 accordingly on a subset of ADNI database was nevertheless obtained not with siamese networks but by majority voting mechanism.

Ortiz-Suárez in \cite{r_1} explored the brain regions most contributing to Alzheimer's disease by applying 2D convolutional neural networks to 2D sMRI brain images (coronal, sagittal and axial cuts). Using the dataset of 85 subjects the authors build a shallow 2D convolutional neural network. Then they create brain models for each filter at the CNN first layer and identify the filters with greatest discriminating power, thus choosing the most contributing brain regions. The authors demonstrated the largest differentiation between patients in the frontal pole region, which is known to host intellectual deficits related to the disease.

\subsubsection{Other networks}

\begin{figure}[h]
\centering\includegraphics[width=0.4\linewidth]{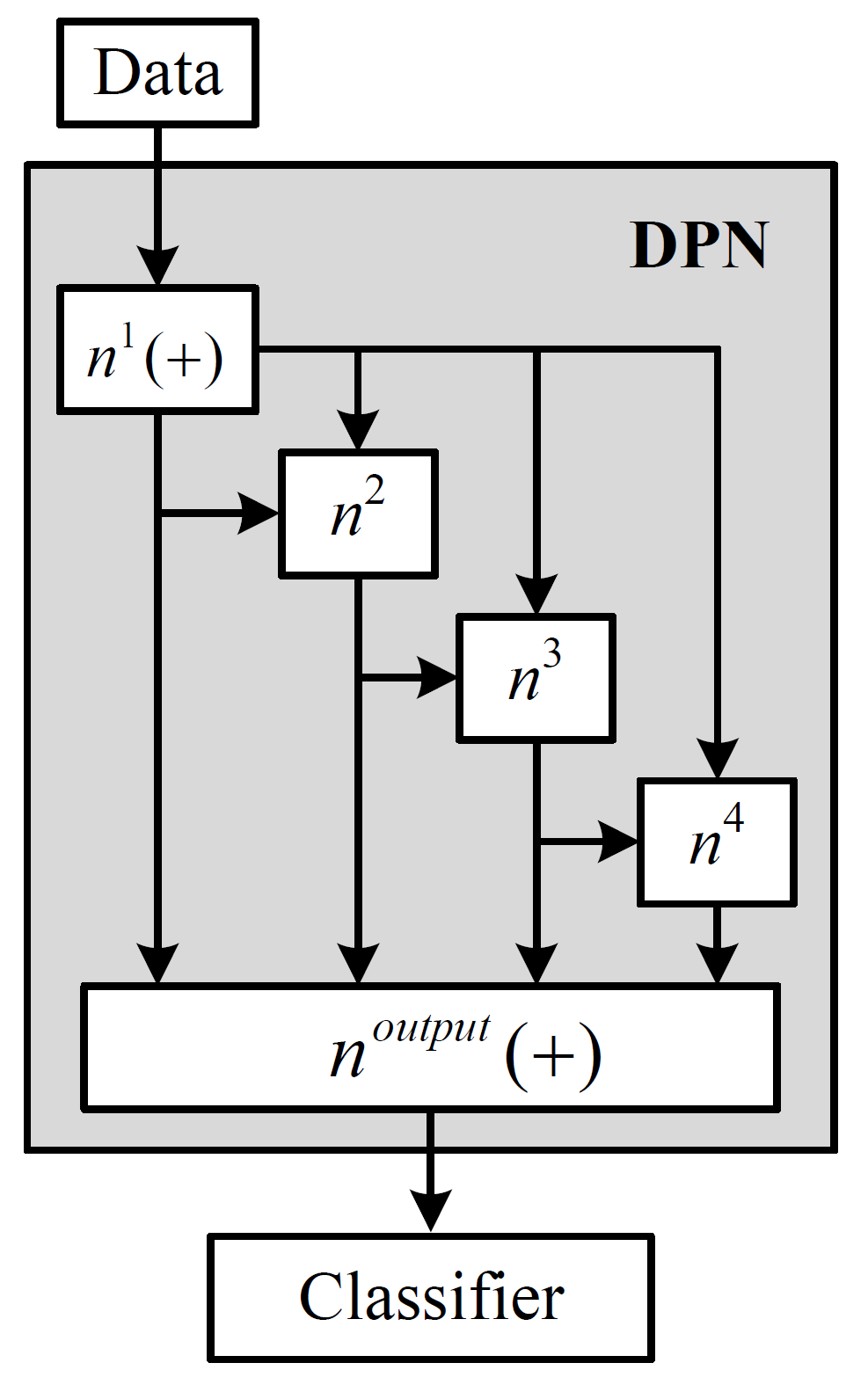}
\caption{An example of a DPN module.}
\label{fig:dpn}
\end{figure}

A new approach was proposed in \cite{c9_DPN_China}. Shi et al. used a deep polinomial network to analyze sMRI and PET images. It differs from classical CNNs by non-linearity of operations.  The building block of the architecture is shown in Fig.\ref{fig:dpn}. Here, \(n^i\) represents a layer of nodes, \((+)\) means a layer of nodes that calculate the weighted sum \(n(z) = \sum_{i}{w_i z_i}\), all other nodes compute \(n(z_1, z_2) = \sum_{i}{w_i (z_1)_i} (z_2)_i\). These blocks were combined into a deep network, the input layers were fed with the average intensity of the 93 ROIs selected on sMRI and PET brain images. A scheme  of the Deep Polynomial Network module is given in figure \ref{fig:dpn} below.

\begin{table*}[t]
\centering
\begin{footnotesize}
\begin{tabular}{|l|l|l|l|l|r|r|r|}
\hline
\multicolumn{1}{|c|}{\multirow{2}{*}{Algorithm}} & \multicolumn{1}{c|}{\multirow{2}{*}{Methodology}} & \multicolumn{1}{c|}{\multirow{2}{*}{Modalities}} & \multicolumn{1}{c|}{\multirow{2}{*}{Content}} & \multicolumn{1}{c|}{\multirow{2}{*}{Data (size)}} & \multicolumn{3}{c|}{Accuracy}                                                          \\ \cline{6-8} 
\multicolumn{1}{|c|}{} & \multicolumn{1}{c|}{} & \multicolumn{1}{c|}{} & \multicolumn{1}{c|}{} & \multicolumn{1}{c|}{} & \multicolumn{1}{c|}{AD/NC} & \multicolumn{1}{c|}{AD/MCI} & \multicolumn{1}{c|}{MCI/NC} \\ \hline

Magnin et al. \cite{volumetric:OBM} & Volumetric & sMRI & Full brain & custom (38) & 0.945 & - & - \\ \hline

Ahmed et al. \cite{f3_Jenny_pcc}  & Feature-based & sMRI & 2 ROIs & ADNI (509) & 0.838 & 0.695 & 0.621 \\ \hline
Ebadi et al. \cite{g1_graph_ensemble_2017} & Feature-based & DTI & Full brain & custom (34) & 0.8 & 0.833 & 0.7 \\ \hline
Lee et al. \cite{g3_svm_DTI_Korea} & Feature-based & DTI & Full brain & LONI (141) & 0.977 & 0.977 & - \\ \hline
Lei et al. \cite{g2_sift_China} & Feature-based & sMRI + PET & Full brain & ADNI (398) & 0.969 & - & 0.866 \\ \hline
Ahmed et al. \cite{f5_Jenny} & Feature-based & sMRI + DTI & 1 ROI & ADNI (203) & 0.902 & 0.766 & 0.794 \\ \hline

Vu et al. \cite{c4_sparse_encoder_Korea} & NN-based & sMRI + PET & Full brain & ADNI (203) & 0.91 & - & - \\ \hline
Payan and Montana \cite{c8_medical_UK} & NN-based & sMRI & Full brain & ADNI (2265) & 0.993 & 1 & 0.942 \\ \hline
Glozman and Liba \cite{c1_report_Standford} & NN-based & sMRI + PET & Full brain & ADNI (1370) & 0.665 & - & - \\ \hline
Sarraf et al. \cite{c7.1_DeepAd_Canada} & NN-based & sMRI, fMRI & Full brain & ADNI (302) & 0.988, 0.999 & - & - \\ \hline
Billones et al. \cite{c5_demnet_Philippines} & NN-based & sMRI & Full brain & ADNI (900) & 0.983 & 0.939 & 0.917 \\ \hline
Aderghal et al. \cite{Aderghal} & NN-based & sMRI & 1 ROI & ADNI (815) & 0.914 & 0.695 & 0.656 \\ \hline
Shi et al. \cite{c9_DPN_China} & NN-based & sMRI + PET & Full brain & ADNI (202) & 0.971 & - & 0.872 \\ \hline
Korolev et al. \cite{c3_Skolkovo} & NN-based & sMRI & Full brain & ADNI (231) & 0.79-0.8 & - & - \\ \hline
Suk et al. \cite{c6_deep_ensemble_sparse_regr_Korea} & NN-based & sMRI & 93 ROIs & ADNI (805) & 0.903 & - & 0.742 \\ \hline
Suk et al. \cite{c6_deep_ensemble_sparse_regr_Korea} & NN-based & sMRI & 93 ROIs & ADNI (805) & 0.903 & - & 0.742 \\ \hline
Luo et al. \cite{ex_2} & NN-based & sMRI & 7 ROIs & ADNI (81) & 0.83 & - & - \\ \hline
Wang et al. \cite{ex_3} & NN-based & sMRI & Full brain & ADNI (629) & - & - & 0.906 \\ \hline
Li et al. \cite{ex_5} & NN-based & sMRI & 1 ROI & ADNI (1776) & 0.965 & 0.67 & 0.622 \\ \hline
Cheng et al. \cite{ex_7} & NN-based & sMRI & 27 ROIs & ADNI (1428) & 0.872 & - & - \\ \hline
Li et al. \cite{ex_9} & NN-based & sMRI & Full brain & ADNI (832) & 0.91 & 0.877 & 0.855 \\ \hline

\end{tabular}
\end{footnotesize}
\label{comparison_table}
\caption{Comparison of different state of the art classification methods.}
\end{table*}

This architecture allowed the authors to reach very good level of accuracy: 0.971, 0.872 for AD/NC, MCI/NC classification. The used algorithm also demonstrated a good level of accuracy (0.789) for MCI-C/MCI-NC determination, where MCI-C  stands for MCI patients that lately converted to AD and MCI-NC stands for MCI patient that were not converted.

In \cite{c3_Skolkovo} the authors compared the residual (ResNet) and plain 3D convolutional neural networks for sMRI image classification. Here the authors examined the four binary classification tasks  AD/LMCI/EMCI/NC, where LMCI and EMCI stands for the late and early MCI stages accordingly. Both networks demonstrated nearly the same performance level, the best figures being obtained for AD/NC classification with  0.79-0.8 accuracies, using 231 sMRI images from a subset of ADNI database.

Residual convolutional networks having shown good performances in computer vision tasks, Li et al. in \cite{ex_5} have also proposed a deep network with residual blocks to preform ordinal ranking.  They compared their model to classical multi-category classification techniques. Data of the only one hippocampal ROI from 1776 sMRI images of ADNI database were used. The final accuracy performance of the proposed method is 0.965, 0.67 and 0.622 for AD/NC, AD/MCI and MCI/NC classification accordingly.

A so-called spectral convolutional neural network was proposed in \cite{ex_9}. It combines classical convolutions with the ability to learn some topological brain features. Li et al. represented a subject's brain as a graph with a set of ROIs as nodes and edges computed using Pearson correlation from a brain grey matter. With a subset of ADNI database containing sMRI images of 832 subjects the authors achieved the classification accuracy 0.91, 0.877, 0.855 for AD/NC, AD/MCI and MCI/NC classification.

In \cite{c6_deep_ensemble_sparse_regr_Korea} Suk et al. try to combine two different methods: sparse regression and convolutional neural networks. The authors got different sparse representations of the 93 ROIs of the sMRI data by varying the sparse control parameter, which allowed them to produce different sets of selected features. Each representation is a vector, so the result of generating multiple representations can be treated as a matrix. This matrix is then fed to the convolutional neural network with 2 convolutional layers and 2 fully-connected layers. This approach led to the classification accuracy level of 0.903 and 0.742 for AD/NC and MCI/NC classification.

\section{Proposed method of classification}
\label{Our method}

\begin{table}[]
\centering
\begin{tabular}{lccc}
\hline
\multicolumn{1}{c}{}                                                           & AD  & MCI & NC  \\ \hline
Subjects                                                                       & 48  & 108 & 58  \\
Samples for train                                                              & 36  & 96  & 46  \\
Samples for test                                                               & 12  & 12  & 12  \\
\begin{tabular}[c]{@{}l@{}}Samples for train\\ after augmentation\end{tabular} & 960 & 960 & 960 \\
\begin{tabular}[c]{@{}l@{}}Samples for test\\ after augmentation\end{tabular}  & 120 & 120 & 120 \\ \hline
\end{tabular}
\caption{Number of patients and data samples for each class before and after augmentation}
\label{table:samples}
\end{table}

\subsection{Data selection}
Data  used  in  the  preparation  of  this  article  were  obtained  from  the  Alzheimer’s  Disease Neuroimaging  Initiative  (ADNI)  database  (\url{http://adni.loni.usc.edu}).
The  ADNI  was  launched  in 2003 as  a  public-private  partnership,  led  by  Principal  Investigator  Michael  W. Weiner, MD. The primary goal of ADNI has been to test whether serial magnetic resonance imaging (MRI),  positron  emission  tomography  (PET),  other  biological  markers,  and  clinical  and neuropsychological  assessment  can  be  combined  to  measure  the  progression  of  mild cognitive impairment (MCI) and early Alzheimer’s disease (AD).
For up-to-date information, see \url{www.adni-info.org}.
We selected 214 subjects: 48 AD patients, 108 MCI and 58 NC (Table \ref{table:samples}). For each patient there is a T1-weighted sMRI image as well as a DTI image. Table \ref{table:demographic} presents a summary of the demographic characteristics of the selected subjects including the age, gender and Mind Mental State Examination (MMSE) score of cognitive functions. In our case the number of images in the dataset is limited by the availability of DTI data. We focus on the hippocampal ROI and surrounding region in the brain scans. 

A preprocess procedure is performed on all used DTI brain images. It includes correction of eddy currents and head motion, skull stripping with Brain Extraction Tool (BET) \cite{BET} and fitting of diffusion tensors to the data with DTIfit module of the FSL software library \cite{FSL}. Fitting step generates MD and FA maps. In the current work we focus only on MD maps of DTI images. To use a normalized anatomical atlas for ROI selection the MD images are affinely co-registered to corresponding sMRI scans. After such co-registration both image modalities are spatially normalized onto the Montreal Neurological Institute (MNI) brain template \cite{MNI}.
Thus, after the preprocessing step, for each patient there is a sMRI and MD-DTI aligned images of the same resolution of \(121 \times 145 \times 121\) voxels.

\begin{table}[]
\centering
\resizebox{\columnwidth}{!}{%
\begin{tabular}{ccccc}
\hline
Diagnosis & Subjects & Age                                                                              & \begin{tabular}[c]{@{}c@{}}Gender\\ (F/M)\end{tabular} & MMSE               \\ \hline
AD        & 48       & \begin{tabular}[c]{@{}c@{}}{[}55.72 - 91.53{]}\\ 75.65 \(\pm\) 8.63\end{tabular} & 20 / 28                                                & 23.0 \(\pm\) 2.42  \\
MCI       & 108      & \begin{tabular}[c]{@{}c@{}}{[}55.32 - 91.88{]}\\ 73.46 \(\pm\) 7.47\end{tabular} & 42 / 66                                                & 27.39 \(\pm\) 1.99 \\
NC        & 58       & \begin{tabular}[c]{@{}c@{}}{[}60.40 - 89.59{]}\\ 73.41 \(\pm\) 5.90\end{tabular} & 30 / 28                                                & 28.88 \(\pm\) 1.18 \\ \hline
\end{tabular}%
}
\caption{Demographic description of the ADNI group, Values are denoted as intervals and as mean \(\pm\) std}
\label{table:demographic}
\end{table}

For the further analysis on each image we select two ROIs (left and write lobes of the hippocampus) as the most discriminative parts of human brain for Alzheimer disease analysis \cite{pierrick}. The ROI selection is performed using atlas AAL \cite{AAL}, the resolution of both hippocampal areas is \(28 \times 28 \times 28 \times 28\) voxels. To compare the influence of the amount of image data passed to the network we also consider the extended ROIs of size \(38 \times 38 \times 38 \times 38\), \(42 \times 42 \times 42 \times 42\) and \(48 \times 48 \times 48 \times 48\) voxels. Herewith, the centers of extended ROIs coincide with the centers of the base ROIs, so the extended ROIs contain all voxels from the base ROI as well as some voxels corresponding to gray matter in the surrounding region.

\subsection{Data augmentation}

We divide the used image database into train and test sets. For test set we select 12 patients from each class, thus leaving 36, 96 and 46 patients for the train set for AD. MCI and NC classes accordingly (Table \ref{table:samples}).

The common problem of using limited dataset for training a neural network is overfitting.
To enlarge the amount of data and prevent the network overfitting we perform data augmentation. The augmentation process is performed separately for the train and test sets. As in many other medical problems the used dataset is imbalanced: the number of patients with MCI is almost 3 times bigger than the number of patients with AD. To eliminate the effect of different class capacities on the network training process we propose to perform a special balancing procedure during data augmentation. Improvements of classification results in case of using data balancing procedure was demonstrated in \cite{r_1}. The main distinctive feature of the method we propose is that class balancing is performed using data augmentation.

The augmentation process with balancing procedure is described with a parameter \(k\), which controls the level of augmentation (i.e. the amount of new images generated from the source ones). Let us suppose the largest set to have \(n_{max}\) instances. In this case all classes should be augmented to \(n_{max} \cdot k\) elements. So for each class that has \(n_i\) elements \(n_{max} \cdot k - n_i\) new elements have to be generated. All new images are generated from the source images by random shift up to 2 pixels in each of the three dimensions $XYZ$ and random Gaussian blur with \(\sigma\) up to \(1.2\).

In this work we have chosen \(k=10\) resulting the number of images to be 960 for each class in training set.
The situation with test dataset is more complicated than with training set. On the one side, the test dataset should not be augmented in order to leave the test data as realistic as possible. On the other side, the size of used test set is very small which leads to a strong discretization of estimation metrics. So, we consider three different test sets: the original set without augmentation (test 0), augmented set using random shifts only (test 1) and augmented set using both random shifts and blur (test 2). Test 0, test 1 and test 2 sets contain 12, 120 and 120 samples accordingly. Thus we can compare this test sets and draw some conclusions about the impact of test set augmentation on the classification results.

\subsection{Network architecture}

\begin{table*}[]
\centering
\resizebox{\textwidth}{!}{%
\begin{tabular}{cccccc}
\hline
\textbf{\begin{tabular}[c]{@{}c@{}}configuration\\ name\end{tabular}} & \textbf{\begin{tabular}[c]{@{}c@{}}number of\\ convolutional\\ layers\end{tabular}} & \textbf{\begin{tabular}[c]{@{}c@{}}convolution kernel\\ size for each layer\end{tabular}} & \textbf{\begin{tabular}[c]{@{}c@{}}number of\\ convolutional filters\\ in each layer\end{tabular}} & \textbf{\begin{tabular}[c]{@{}c@{}}number of\\ fully-connected\\ layers\end{tabular}} & \textbf{\begin{tabular}[c]{@{}c@{}}number of units in\\  each fully-connected\\  layer\end{tabular}} \\ \hline
C1 & 4 & (5, 4, 3, 3) & (16, 32, 64, 128) & 2 & (16, 8) \\
C2 & 5 & (5, 4, 3, 3, 3) & (16, 32, 64, 128, 128) & 2 & (16, 8) \\
C3 & 5 & (7, 6, 5, 4, 3) & (16, 32, 64, 128, 256) & 2 & (32, 8) \\
C4 & 6 & (7, 6, 5, 4, 3, 3) & (16, 32, 64, 128, 256, 256) & 1 & 16 \\ \hline
\end{tabular}%
}
\caption{Compared architectures of the used neural networks}
\label{table:architectures}
\end{table*}

\begin{figure}[h]
\centering\includegraphics[width=0.6\linewidth]{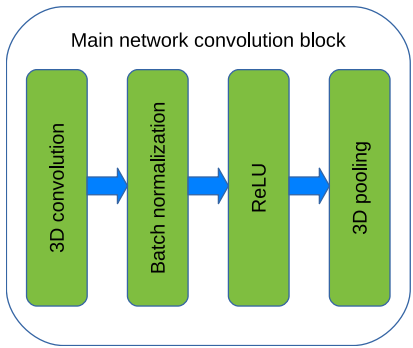}
\caption{Main convolutional block of the proposed network architecture.}
\label{fig:NN_block}
\end{figure}

In this work we use a number of 3D convolutional neural networks with slightly different configuration and compare them. The base building block of the used networks consists of 4 consistent operations: 3D convolution, batch normalization \cite{batch_norm_2015}, applying rectifier linear unit and 3D pooling illustrated in Figure \ref{fig:NN_block}.

\begin{figure}[h]
\centering\includegraphics[width=\linewidth]{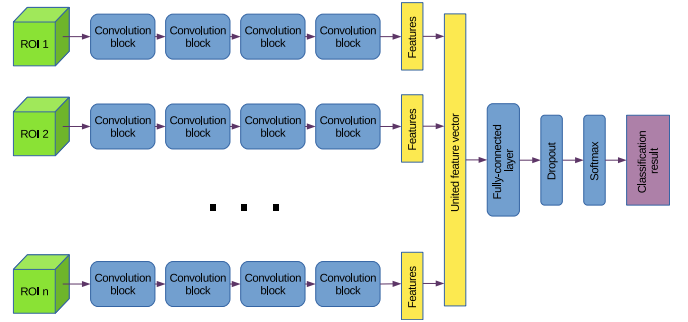}
\caption{Proposed convolutional neural network architecture.}
\label{fig:NN_architecture}
\end{figure}

For each ROI in the image and for each modality we use a separate pipeline of the described blocks. Each pipeline ends with a flatten operation, after which the outputs of the pipelines are concatenated and are passed to the fully-connected layer. This fully-connected layer is followed with a dropout layer and the softmax layer, which produces the network output (Figure \ref{fig:NN_architecture}). Thus the described network is a siamese network which performs the late fusion of the data from input ROIs.

The usage of batch normalization allows us to speed up the network training process and according to \cite{batch_norm_2015} eliminate the necessity of using the pretraining techniques (e.g. autoencoders). Batch normalization partially plays a role of regularization and also allows each layer of a network to learn by itself a little bit more independently of other layers.

The network training process is performed by minimizing the euclidean loss function with Nesterov momentum optimization:
\[v_{t+1}=m v_t - \mu_t \nabla(w_t + m v_t),\]
\[w_{t+1} = w_t + v_{t+1}, \]
\[v_0 = 0,\]
where \(v_t\) is the velocity at each iteration \(t\) and \(w_t\) is the optimized vector of network weights at iteration \(t\), \(m\) is the momentum value, \(\mu_t\) is the learning rate. The momentum value we use is \(m=0.93\). 
The learning rate is updated exponentially:
\[\mu_{t+1} = \mu_t \lambda^{t/t_0},\]
where \(\lambda\) is a decay rate and \(t_0\) is a decay step. Here the division \(t/t_0\) is an integer division, so the learning rate updates ones every \(t_0\) iterations. We use decay rate \(\lambda=0.8\), decay step \(t_0=100\) and the initial value of learning rate \(\mu_0=0.01\).These values were chosen empirically in accordance with our preliminary experiments yielded the best accuracies. 

In this work we compare a few convolutional neural networks differing in their parameters such as a number of convolutional layers and convolutional filters (Table \ref{table:architectures}). We intentionally consider architectures with different depth as we use ROIs with different sizes. Specifically, we use more shallow C1 and C2 architectures for ROIs of size 28 and 38, and deeper C3 and C4 architectures for ROIs of size 42 and 48.

To prevent the network overfitting we use a method similar to a 10-fold cross-validation. During training we randomly select 90\% of training data and use it to learn the network weights, while 10\% left data is used as validation. After the fixed number of iterations this train-test separation is repeated again. This approach leads to the effective usage of the available training data.

For more accurate comparison of the networks we use the same number \(q=90\) of batches passed into every of the \(n\) inputs. If the whole number of batches does not fit into the memory, we separate them into a several mini-groups and correct the weights of the network after accumulating the gradients of all mini-groups. Additionally each network is trained with 1000 iterations.

\section{Results}
\label{sec:Results}
\begin{table*}[]
\centering
\resizebox{\textwidth}{!}{%
\begin{tabular}{c|cc|ccc|ccc|ccc}
\hline
\rowcolor[HTML]{EFEFEF} 
\cellcolor[HTML]{EFEFEF} & \cellcolor[HTML]{EFEFEF} & \cellcolor[HTML]{EFEFEF} & \multicolumn{3}{c|}{\cellcolor[HTML]{EFEFEF}\textbf{\begin{tabular}[c]{@{}c@{}}Top-mean ACC\\ {[}95\% CI{]}\end{tabular}}} & \multicolumn{3}{c|}{\cellcolor[HTML]{EFEFEF}\textbf{\begin{tabular}[c]{@{}c@{}}Top-mean SEN\\ {[}95\% CI{]}\end{tabular}}} & \multicolumn{3}{c}{\cellcolor[HTML]{EFEFEF}\textbf{\begin{tabular}[c]{@{}c@{}}Top-mean SPC\\ {[}95\% CI{]}\end{tabular}}} \\
\rowcolor[HTML]{EFEFEF} 
\multirow{-2}{*}{\cellcolor[HTML]{EFEFEF}\textbf{\begin{tabular}[c]{@{}c@{}}used\\ data\end{tabular}}} & \multirow{-2}{*}{\cellcolor[HTML]{EFEFEF}\textbf{\begin{tabular}[c]{@{}c@{}}ROI\\ size\end{tabular}}} & \multirow{-2}{*}{\cellcolor[HTML]{EFEFEF}\textbf{\begin{tabular}[c]{@{}c@{}}configu-\\ ration\end{tabular}}} & \textbf{test 0} & \textbf{test 1} & \textbf{test 2} & \textbf{test 0} & \textbf{test 1} & \textbf{test 2} & \textbf{test 0} & \textbf{test 1} & \textbf{test 2} \\ \hline
 &  & C1 & \textbf{\begin{tabular}[c]{@{}c@{}}0.854\\ \(\pm\) 0.141\end{tabular}} & \begin{tabular}[c]{@{}c@{}}0.827\\ \(\pm\) 0.048\end{tabular} & \textbf{\begin{tabular}[c]{@{}c@{}}0.839\\ \(\pm\) 0.047\end{tabular}} & \begin{tabular}[c]{@{}c@{}}0.883\\ \(\pm\) 0.123\end{tabular} & \begin{tabular}[c]{@{}c@{}}0.880\\ \(\pm\) 0.041\end{tabular} & \begin{tabular}[c]{@{}c@{}}0.883\\ \(\pm\) 0.041\end{tabular} & \begin{tabular}[c]{@{}c@{}}0.900\\ \(\pm\) 0.110\end{tabular} & \textbf{\begin{tabular}[c]{@{}c@{}}0.885\\ \(\pm\) 0.040\end{tabular}} & \textbf{\begin{tabular}[c]{@{}c@{}}0.901\\ \(\pm\) 0.038\end{tabular}} \\
 & \multirow{-2}{*}{28} & \cellcolor[HTML]{EFEFEF}C2 & \cellcolor[HTML]{EFEFEF}\begin{tabular}[c]{@{}c@{}}0.825\\ \(\pm\)0.152\end{tabular} & \cellcolor[HTML]{EFEFEF}\begin{tabular}[c]{@{}c@{}}0.813\\ \(\pm\)0.049\end{tabular} & \cellcolor[HTML]{EFEFEF}\begin{tabular}[c]{@{}c@{}}0.813\\ \(\pm\)0.049\end{tabular} & \cellcolor[HTML]{EFEFEF}\begin{tabular}[c]{@{}c@{}}0.867\\ \(\pm\)0.135\end{tabular} & \cellcolor[HTML]{EFEFEF}\begin{tabular}[c]{@{}c@{}}0.848\\ \(\pm\)0.046\end{tabular} & \cellcolor[HTML]{EFEFEF}\begin{tabular}[c]{@{}c@{}}0.850\\ \(\pm\)0.045\end{tabular} & \cellcolor[HTML]{EFEFEF}\textbf{\begin{tabular}[c]{@{}c@{}}0.908\\ \(\pm\)0.104\end{tabular}} & \cellcolor[HTML]{EFEFEF}\begin{tabular}[c]{@{}c@{}}0.879\\ \(\pm\)0.0412\end{tabular} & \cellcolor[HTML]{EFEFEF}\begin{tabular}[c]{@{}c@{}}0.869\\ \(\pm\)0.043\end{tabular} \\
 & \cellcolor[HTML]{EFEFEF} & C1 & \begin{tabular}[c]{@{}c@{}}0.808\\ \(\pm\)0.158\end{tabular} & \begin{tabular}[c]{@{}c@{}}0.790\\ \(\pm\)0.052\end{tabular} & \begin{tabular}[c]{@{}c@{}}0.801\\ \(\pm\)0.051\end{tabular} & \begin{tabular}[c]{@{}c@{}}0.775\\ \(\pm\)0.167\end{tabular} & \begin{tabular}[c]{@{}c@{}}0.799\\ \(\pm\)0.051\end{tabular} & \begin{tabular}[c]{@{}c@{}}0.786\\ \(\pm\)0.052\end{tabular} & \begin{tabular}[c]{@{}c@{}}0.883\\ \(\pm\)0.123\end{tabular} & \begin{tabular}[c]{@{}c@{}}0.831\\ \(\pm\)0.047\end{tabular} & \begin{tabular}[c]{@{}c@{}}0.863\\ \(\pm\)0.044\end{tabular} \\
 & \multirow{-2}{*}{\cellcolor[HTML]{EFEFEF}38} & \cellcolor[HTML]{EFEFEF}C2 & \cellcolor[HTML]{EFEFEF}\begin{tabular}[c]{@{}c@{}}0.767\\ \(\pm\)0.169\end{tabular} & \cellcolor[HTML]{EFEFEF}\begin{tabular}[c]{@{}c@{}}0.769\\ \(\pm\)0.053\end{tabular} & \cellcolor[HTML]{EFEFEF}\begin{tabular}[c]{@{}c@{}}0.776\\ \(\pm\)0.053\end{tabular} & \cellcolor[HTML]{EFEFEF}\begin{tabular}[c]{@{}c@{}}0.825\\ \(\pm\)0.152\end{tabular} & \cellcolor[HTML]{EFEFEF}\begin{tabular}[c]{@{}c@{}}0.809\\ \(\pm\)0.050\end{tabular} & \cellcolor[HTML]{EFEFEF}\begin{tabular}[c]{@{}c@{}}0.804\\ \(\pm\)0.050\end{tabular} & \cellcolor[HTML]{EFEFEF}\begin{tabular}[c]{@{}c@{}}0.817\\ \(\pm\)0.155\end{tabular} & \cellcolor[HTML]{EFEFEF}\begin{tabular}[c]{@{}c@{}}0.816\\ \(\pm\)0.049\end{tabular} & \cellcolor[HTML]{EFEFEF}\begin{tabular}[c]{@{}c@{}}0.828\\ \(\pm\)0.048\end{tabular} \\
\multirow{-4}{*}{\begin{tabular}[c]{@{}c@{}}sMRI\_L\\ +\\ sMRI\_R\end{tabular}} &  & C3 & \begin{tabular}[c]{@{}c@{}}0.800\\ \(\pm\)0.160\end{tabular} & \begin{tabular}[c]{@{}c@{}}0.777\\ \(\pm\)0.053\end{tabular} & \begin{tabular}[c]{@{}c@{}}0.790\\ \(\pm\)0.052\end{tabular} & \begin{tabular}[c]{@{}c@{}}0.792\\ \(\pm\)0.163\end{tabular} & \begin{tabular}[c]{@{}c@{}}0.813\\ \(\pm\)0.049\end{tabular} & \begin{tabular}[c]{@{}c@{}}0.790\\ \(\pm\)0.052\end{tabular} & \begin{tabular}[c]{@{}c@{}}0.883\\ \(\pm\)0.123\end{tabular} & \begin{tabular}[c]{@{}c@{}}0.894\\ \(\pm\)0.039\end{tabular} & \begin{tabular}[c]{@{}c@{}}0.882\\ \(\pm\)0.041\end{tabular} \\
 & \multirow{-2}{*}{42} & \cellcolor[HTML]{EFEFEF}C4 & \cellcolor[HTML]{EFEFEF}\begin{tabular}[c]{@{}c@{}}0.771\\ \(\pm\)0.168\end{tabular} & \cellcolor[HTML]{EFEFEF}\begin{tabular}[c]{@{}c@{}}0.725\\ \(\pm\)0.057\end{tabular} & \cellcolor[HTML]{EFEFEF}\begin{tabular}[c]{@{}c@{}}0.757\\ \(\pm\)0.054\end{tabular} & \cellcolor[HTML]{EFEFEF}\begin{tabular}[c]{@{}c@{}}0.725\\ \(\pm\)0.179\end{tabular} & \cellcolor[HTML]{EFEFEF}\begin{tabular}[c]{@{}c@{}}0.711\\ \(\pm\)0.057\end{tabular} & \cellcolor[HTML]{EFEFEF}\begin{tabular}[c]{@{}c@{}}0.724\\ \(\pm\)0.057\end{tabular} & \cellcolor[HTML]{EFEFEF}\begin{tabular}[c]{@{}c@{}}0.892\\ \(\pm\)0.116\end{tabular} & \cellcolor[HTML]{EFEFEF}\begin{tabular}[c]{@{}c@{}}0.821\\ \(\pm\)0.049\end{tabular} & \cellcolor[HTML]{EFEFEF}\begin{tabular}[c]{@{}c@{}}0.827\\ \(\pm\)0.048\end{tabular} \\
 & \cellcolor[HTML]{EFEFEF} & C3 & \begin{tabular}[c]{@{}c@{}}0.788\\ \(\pm\)0.164\end{tabular} & \begin{tabular}[c]{@{}c@{}}0.779\\ \(\pm\)0.053\end{tabular} & \begin{tabular}[c]{@{}c@{}}0.779\\ \(\pm\)0.053\end{tabular} & \begin{tabular}[c]{@{}c@{}}0.767\\ \(\pm\)0.169\end{tabular} & \begin{tabular}[c]{@{}c@{}}0.742\\ \(\pm\)0.055\end{tabular} & \begin{tabular}[c]{@{}c@{}}0.738\\ \(\pm\)0.056\end{tabular} & \begin{tabular}[c]{@{}c@{}}0.883\\ \(\pm\)0.123\end{tabular} & \begin{tabular}[c]{@{}c@{}}0.864\\ \(\pm\)0.043\end{tabular} & \begin{tabular}[c]{@{}c@{}}0.854\\ \(\pm\)0.045\end{tabular} \\
 & \multirow{-2}{*}{\cellcolor[HTML]{EFEFEF}48} & \cellcolor[HTML]{EFEFEF}C4 & \cellcolor[HTML]{EFEFEF}\begin{tabular}[c]{@{}c@{}}0.842\\ \(\pm\)0.146\end{tabular} & \cellcolor[HTML]{EFEFEF}\textbf{\begin{tabular}[c]{@{}c@{}}0.829\\ \(\pm\)0.048\end{tabular}} & \cellcolor[HTML]{EFEFEF}\begin{tabular}[c]{@{}c@{}}0.836\\ \(\pm\)0.047\end{tabular} & \cellcolor[HTML]{EFEFEF}\textbf{\begin{tabular}[c]{@{}c@{}}0.925\\ \(\pm\)0.090\end{tabular}} & \cellcolor[HTML]{EFEFEF}\textbf{\begin{tabular}[c]{@{}c@{}}0.888\\ \(\pm\)0.040\end{tabular}} & \cellcolor[HTML]{EFEFEF}\textbf{\begin{tabular}[c]{@{}c@{}}0.898\\ \(\pm\)0.038\end{tabular}} & \cellcolor[HTML]{EFEFEF}\begin{tabular}[c]{@{}c@{}}0.858\\ \(\pm\)0.140\end{tabular} & \cellcolor[HTML]{EFEFEF}\begin{tabular}[c]{@{}c@{}}0.852\\ \(\pm\)0.045\end{tabular} & \cellcolor[HTML]{EFEFEF}\begin{tabular}[c]{@{}c@{}}0.838\\ \(\pm\)0.047\end{tabular} \\ \hline
 &  & C1 & \begin{tabular}[c]{@{}c@{}}0.875\\ \(\pm\) 0.129\end{tabular} & \begin{tabular}[c]{@{}c@{}}0.874\\ \(\pm\) 0.042\end{tabular} & \begin{tabular}[c]{@{}c@{}}0.854\\ \(\pm\)0.045\end{tabular} & \begin{tabular}[c]{@{}c@{}}0.892\\ \(\pm\)0.116\end{tabular} & \begin{tabular}[c]{@{}c@{}}0.901\\ \(\pm\)0.038\end{tabular} & \begin{tabular}[c]{@{}c@{}}0.874\\ \(\pm\)0.042\end{tabular} & \begin{tabular}[c]{@{}c@{}}0.892\\ \(\pm\)0.116\end{tabular} & \begin{tabular}[c]{@{}c@{}}0.901\\ \(\pm\)0.038\end{tabular} & \begin{tabular}[c]{@{}c@{}}0.887\\ \(\pm\)0.040\end{tabular} \\
 & \multirow{-2}{*}{28} & \cellcolor[HTML]{EFEFEF}C2 & \cellcolor[HTML]{EFEFEF}\begin{tabular}[c]{@{}c@{}}0.892\\ \(\pm\)0.116\end{tabular} & \cellcolor[HTML]{EFEFEF}\begin{tabular}[c]{@{}c@{}}0.904\\ \(\pm\)0.037\end{tabular} & \cellcolor[HTML]{EFEFEF}\begin{tabular}[c]{@{}c@{}}0.885\\ \(\pm\)0.040\end{tabular} & \cellcolor[HTML]{EFEFEF}\begin{tabular}[c]{@{}c@{}}0.933\\ \(\pm\)0.083\end{tabular} & \cellcolor[HTML]{EFEFEF}\begin{tabular}[c]{@{}c@{}}0.940\\ \(\pm\)0.030\end{tabular} & \cellcolor[HTML]{EFEFEF}\begin{tabular}[c]{@{}c@{}}0.904\\ \(\pm\)0.037\end{tabular} & \cellcolor[HTML]{EFEFEF}\begin{tabular}[c]{@{}c@{}}0.917\\ \(\pm\)0.097\end{tabular} & \cellcolor[HTML]{EFEFEF}\begin{tabular}[c]{@{}c@{}}0.914\\ \(\pm\)0.035\end{tabular} & \cellcolor[HTML]{EFEFEF}\begin{tabular}[c]{@{}c@{}}0.912\\ \(\pm\)0.036\end{tabular} \\
 & \cellcolor[HTML]{EFEFEF} & C1 & \begin{tabular}[c]{@{}c@{}}0.883\\ \(\pm\)0.123\end{tabular} & \begin{tabular}[c]{@{}c@{}}0.904\\ \(\pm\)0.037\end{tabular} & \begin{tabular}[c]{@{}c@{}}0.891\\ \(\pm\)0.040\end{tabular} & \begin{tabular}[c]{@{}c@{}}0.925\\ \(\pm\)0.090\end{tabular} & \begin{tabular}[c]{@{}c@{}}0.923\\ \(\pm\)0.034\end{tabular} & \begin{tabular}[c]{@{}c@{}}0.918\\ \(\pm\)0.035\end{tabular} & \begin{tabular}[c]{@{}c@{}}0.883\\ \(\pm\)0.123\end{tabular} & \begin{tabular}[c]{@{}c@{}}0.910\\ \(\pm\)0.036\end{tabular} & \begin{tabular}[c]{@{}c@{}}0.910\\ \(\pm\)0.036\end{tabular} \\
\multirow{-3}{*}{\begin{tabular}[c]{@{}c@{}}sMRI\_L\\ +\\ sMRI\_R\\ +\\ DTI\_L\\ +\\ DTI\_R\end{tabular}} & \multirow{-2}{*}{\cellcolor[HTML]{EFEFEF}38} & \cellcolor[HTML]{EFEFEF}C2 & \cellcolor[HTML]{EFEFEF}\begin{tabular}[c]{@{}c@{}}0.842\\ \(\pm\)0.146\end{tabular} & \cellcolor[HTML]{EFEFEF}\begin{tabular}[c]{@{}c@{}}0.854\\ \(\pm\)0.045\end{tabular} & \cellcolor[HTML]{EFEFEF}\begin{tabular}[c]{@{}c@{}}0.842\\ \(\pm\)0.046\end{tabular} & \cellcolor[HTML]{EFEFEF}\begin{tabular}[c]{@{}c@{}}0.925\\ \(\pm\)0.090\end{tabular} & \cellcolor[HTML]{EFEFEF}\begin{tabular}[c]{@{}c@{}}0.873\\ \(\pm\)0.042\end{tabular} & \cellcolor[HTML]{EFEFEF}\begin{tabular}[c]{@{}c@{}}0.892\\ \(\pm\)0.039\end{tabular} & \cellcolor[HTML]{EFEFEF}\begin{tabular}[c]{@{}c@{}}0.875\\ \(\pm\)0.129\end{tabular} & \cellcolor[HTML]{EFEFEF}\begin{tabular}[c]{@{}c@{}}0.883\\ \(\pm\)0.041\end{tabular} & \cellcolor[HTML]{EFEFEF}\begin{tabular}[c]{@{}c@{}}0.872\\ \(\pm\)0.042\end{tabular} \\
 &  & C3 & \begin{tabular}[c]{@{}c@{}}0.900\\ \(\pm\)0.110\end{tabular} & \begin{tabular}[c]{@{}c@{}}0.908\\ \(\pm\)0.037\end{tabular} & \begin{tabular}[c]{@{}c@{}}0.895\\ \(\pm\)0.039\end{tabular} & \begin{tabular}[c]{@{}c@{}}0.933\\ \(\pm\)0.083\end{tabular} & \begin{tabular}[c]{@{}c@{}}0.907\\ \(\pm\)0.037\end{tabular} & \begin{tabular}[c]{@{}c@{}}0.910\\ \(\pm\)0.036\end{tabular} & \begin{tabular}[c]{@{}c@{}}0.917\\ \(\pm\)0.097\end{tabular} & \begin{tabular}[c]{@{}c@{}}0.924\\ \(\pm\)0.034\end{tabular} & \begin{tabular}[c]{@{}c@{}}0.913\\ \(\pm\)0.036\end{tabular} \\
 & \multirow{-2}{*}{42} & \cellcolor[HTML]{EFEFEF}C4 & \cellcolor[HTML]{EFEFEF}\begin{tabular}[c]{@{}c@{}}0.908\\ \(\pm\)0.104\end{tabular} & \cellcolor[HTML]{EFEFEF}\begin{tabular}[c]{@{}c@{}}0.871\\ \(\pm\)0.042\end{tabular} & \cellcolor[HTML]{EFEFEF}\begin{tabular}[c]{@{}c@{}}0.866\\ \(\pm\)0.043\end{tabular} & \cellcolor[HTML]{EFEFEF}\begin{tabular}[c]{@{}c@{}}0.850\\ \(\pm\)0.143\end{tabular} & \cellcolor[HTML]{EFEFEF}\begin{tabular}[c]{@{}c@{}}0.823\\ \(\pm\)0.048\end{tabular} & \cellcolor[HTML]{EFEFEF}\begin{tabular}[c]{@{}c@{}}0.806\\ \(\pm\)0.050\end{tabular} & \cellcolor[HTML]{EFEFEF}\begin{tabular}[c]{@{}c@{}}0.992\\ \(\pm\)0.022\end{tabular} & \cellcolor[HTML]{EFEFEF}\begin{tabular}[c]{@{}c@{}}0.973\\ \(\pm\)0.021\end{tabular} & \cellcolor[HTML]{EFEFEF}\begin{tabular}[c]{@{}c@{}}0.956\\ \(\pm\)0.026\end{tabular} \\
 & \cellcolor[HTML]{EFEFEF} & C3 & \begin{tabular}[c]{@{}c@{}}0.904\\ \(\pm\)0.107\end{tabular} & \begin{tabular}[c]{@{}c@{}}0.903\\ \(\pm\)0.038\end{tabular} & \begin{tabular}[c]{@{}c@{}}0.895\\ \(\pm\)0.039\end{tabular} & \begin{tabular}[c]{@{}c@{}}0.867\\ \(\pm\)0.135\end{tabular} & \begin{tabular}[c]{@{}c@{}}0.883\\ \(\pm\)0.041\end{tabular} & \begin{tabular}[c]{@{}c@{}}0.871\\ \(\pm\)0.042\end{tabular} & \begin{tabular}[c]{@{}c@{}}0.942\\ \(\pm\)0.076\end{tabular} & \begin{tabular}[c]{@{}c@{}}0.923\\ \(\pm\)0.034\end{tabular} & \begin{tabular}[c]{@{}c@{}}0.924\\ \(\pm\)0.034\end{tabular} \\
 & \multirow{-2}{*}{\cellcolor[HTML]{EFEFEF}48} & \cellcolor[HTML]{EFEFEF}C4 & \cellcolor[HTML]{EFEFEF}\textbf{\begin{tabular}[c]{@{}c@{}}0.967\\ \(\pm\) 0.055\end{tabular}} & \cellcolor[HTML]{EFEFEF}\textbf{\begin{tabular}[c]{@{}c@{}}0.938\\ \(\pm\) 0.03\end{tabular}} & \cellcolor[HTML]{EFEFEF}\textbf{\begin{tabular}[c]{@{}c@{}}0.947\\ \(\pm\) 0.03\end{tabular}} & \cellcolor[HTML]{EFEFEF}\textbf{\begin{tabular}[c]{@{}c@{}}0.958\\ \(\pm\) 0.06\end{tabular}} & \cellcolor[HTML]{EFEFEF}\textbf{\begin{tabular}[c]{@{}c@{}}0.943\\ \(\pm\) 0.03\end{tabular}} & \cellcolor[HTML]{EFEFEF}\textbf{\begin{tabular}[c]{@{}c@{}}0.936\\ \(\pm\) 0.035\end{tabular}} & \cellcolor[HTML]{EFEFEF}\textbf{\begin{tabular}[c]{@{}c@{}}0.975\\ \(\pm\) 0.06\end{tabular}} & \cellcolor[HTML]{EFEFEF}\textbf{\begin{tabular}[c]{@{}c@{}}0.939\\ \(\pm\) 0.03\end{tabular}} & \cellcolor[HTML]{EFEFEF}\textbf{\begin{tabular}[c]{@{}c@{}}0.964\\ \(\pm\) 0.035\end{tabular}} \\ \hline
\end{tabular}%
}\caption{AD-NC classification results. Here \(\alpha \pm \beta\) designation corresponds to the top-mean value of metric and it's \(95\%\) confidence interval. \(\alpha\) - metric value, \([\alpha-\beta, \alpha+\beta]\) - confidence interval.}
\label{table:results_AD_NC}
\end{table*}
\begin{table*}[]
\centering
\resizebox{\textwidth}{!}{%
\begin{tabular}{c|cc|ccc|ccc|ccc}
\hline
\rowcolor[HTML]{EFEFEF} 
\cellcolor[HTML]{EFEFEF} & \cellcolor[HTML]{EFEFEF} & \cellcolor[HTML]{EFEFEF} & \multicolumn{3}{c|}{\cellcolor[HTML]{EFEFEF}\textbf{\begin{tabular}[c]{@{}c@{}}Top-mean ACC\\ {[}95\% CI{]}\end{tabular}}} & \multicolumn{3}{c|}{\cellcolor[HTML]{EFEFEF}\textbf{\begin{tabular}[c]{@{}c@{}}Top-mean SEN\\ {[}95\% CI{]}\end{tabular}}} & \multicolumn{3}{c}{\cellcolor[HTML]{EFEFEF}\textbf{\begin{tabular}[c]{@{}c@{}}Top-mean SPC\\ {[}95\% CI{]}\end{tabular}}} \\
\rowcolor[HTML]{EFEFEF} 
\multirow{-2}{*}{\cellcolor[HTML]{EFEFEF}\textbf{\begin{tabular}[c]{@{}c@{}}used\\ data\end{tabular}}} & \multirow{-2}{*}{\cellcolor[HTML]{EFEFEF}\textbf{\begin{tabular}[c]{@{}c@{}}ROI\\ size\end{tabular}}} & \multirow{-2}{*}{\cellcolor[HTML]{EFEFEF}\textbf{\begin{tabular}[c]{@{}c@{}}configu-\\ ration\end{tabular}}} & \textbf{test 0} & \textbf{test 1} & \textbf{test 2} & \textbf{test 0} & \textbf{test 1} & \textbf{test 2} & \textbf{test 0} & \textbf{test 1} & \textbf{test 2} \\ \hline
 &  & C1 & \textbf{\begin{tabular}[c]{@{}c@{}}0.754\\ \(\pm\)0.172\end{tabular}} & \textbf{\begin{tabular}[c]{@{}c@{}}0.760\\ \(\pm\)0.054\end{tabular}} & \textbf{\begin{tabular}[c]{@{}c@{}}0.759\\ \(\pm\)0.054\end{tabular}} & \begin{tabular}[c]{@{}c@{}}0.617\\ \(\pm\)0.195\end{tabular} & \begin{tabular}[c]{@{}c@{}}0.638\\ \(\pm\)0.061\end{tabular} & \begin{tabular}[c]{@{}c@{}}0.639\\ \(\pm\)0.061\end{tabular} & \begin{tabular}[c]{@{}c@{}}0.950\\ \(\pm\)0.069\end{tabular} & \begin{tabular}[c]{@{}c@{}}0.962\\ \(\pm\)0.024\end{tabular} & \begin{tabular}[c]{@{}c@{}}0.953\\ \(\pm\)0.027\end{tabular} \\
 & \multirow{-2}{*}{28} & \cellcolor[HTML]{EFEFEF}C2 & \cellcolor[HTML]{EFEFEF}\begin{tabular}[c]{@{}c@{}}0.746\\ \(\pm\)0.174\end{tabular} & \cellcolor[HTML]{EFEFEF}\begin{tabular}[c]{@{}c@{}}0.736\\ \(\pm\)0.056\end{tabular} & \cellcolor[HTML]{EFEFEF}\begin{tabular}[c]{@{}c@{}}0.754\\ \(\pm\)0.054\end{tabular} & \cellcolor[HTML]{EFEFEF}\begin{tabular}[c]{@{}c@{}}0.600\\ \(\pm\)0.196\end{tabular} & \cellcolor[HTML]{EFEFEF}\begin{tabular}[c]{@{}c@{}}0.633\\ \(\pm\)0.061\end{tabular} & \cellcolor[HTML]{EFEFEF}\begin{tabular}[c]{@{}c@{}}0.609\\ \(\pm\)0.062\end{tabular} & \cellcolor[HTML]{EFEFEF}\begin{tabular}[c]{@{}c@{}}0.967\\ \(\pm\)0.053\end{tabular} & \cellcolor[HTML]{EFEFEF}\begin{tabular}[c]{@{}c@{}}0.959\\ \(\pm\)0.025\end{tabular} & \cellcolor[HTML]{EFEFEF}\begin{tabular}[c]{@{}c@{}}0.956\\ \(\pm\)0.026\end{tabular} \\
 & \cellcolor[HTML]{EFEFEF} & C1 & \begin{tabular}[c]{@{}c@{}}0.642\\ \(\pm\)0.192\end{tabular} & \begin{tabular}[c]{@{}c@{}}0.677\\ \(\pm\)0.059\end{tabular} & \begin{tabular}[c]{@{}c@{}}0.663\\ \(\pm\)0.060\end{tabular} & \begin{tabular}[c]{@{}c@{}}0.525\\ \(\pm\)0.200\end{tabular} & \begin{tabular}[c]{@{}c@{}}0.565\\ \(\pm\)0.063\end{tabular} & \begin{tabular}[c]{@{}c@{}}0.526\\ \(\pm\)0.063\end{tabular} & \textbf{\begin{tabular}[c]{@{}c@{}}1.00\\ \(\pm\)0.000\end{tabular}} & \begin{tabular}[c]{@{}c@{}}0.982\\ \(\pm\)0.017\end{tabular} & \begin{tabular}[c]{@{}c@{}}0.958\\ \(\pm\)0.025\end{tabular} \\
\multirow{-3}{*}{\begin{tabular}[c]{@{}c@{}}sMRI\_L\\ +\\ sMRI\_R\end{tabular}} & \multirow{-2}{*}{\cellcolor[HTML]{EFEFEF}38} & \cellcolor[HTML]{EFEFEF}C2 & \cellcolor[HTML]{EFEFEF}\begin{tabular}[c]{@{}c@{}}0.679\\ \(\pm\)0.187\end{tabular} & \cellcolor[HTML]{EFEFEF}\begin{tabular}[c]{@{}c@{}}0.731\\ \(\pm\)0.056\end{tabular} & \cellcolor[HTML]{EFEFEF}\begin{tabular}[c]{@{}c@{}}0.690\\ \(\pm\)0.059\end{tabular} & \cellcolor[HTML]{EFEFEF}\textbf{\begin{tabular}[c]{@{}c@{}}0.658\\ \(\pm\)0.190\end{tabular}} & \cellcolor[HTML]{EFEFEF}\textbf{\begin{tabular}[c]{@{}c@{}}0.708\\ \(\pm\)0.058\end{tabular}} & \cellcolor[HTML]{EFEFEF}\textbf{\begin{tabular}[c]{@{}c@{}}0.711\\ \(\pm\)0.057\end{tabular}} & \cellcolor[HTML]{EFEFEF}\begin{tabular}[c]{@{}c@{}}0.992\\ \(\pm\)0.022\end{tabular} & \cellcolor[HTML]{EFEFEF}\textbf{\begin{tabular}[c]{@{}c@{}}0.999\\ \(\pm\)0.002\end{tabular}} & \cellcolor[HTML]{EFEFEF}\begin{tabular}[c]{@{}c@{}}0.991\\ \(\pm\)0.011\end{tabular} \\
 &  & C3 & \begin{tabular}[c]{@{}c@{}}0.688\\ \(\pm\)0.185\end{tabular} & \begin{tabular}[c]{@{}c@{}}0.651\\ \(\pm\)0.060\end{tabular} & \begin{tabular}[c]{@{}c@{}}0.678\\ \(\pm\)0.059\end{tabular} & \begin{tabular}[c]{@{}c@{}}0.592\\ \(\pm\)0.197\end{tabular} & \begin{tabular}[c]{@{}c@{}}0.553\\ \(\pm\)0.063\end{tabular} & \begin{tabular}[c]{@{}c@{}}0.610\\ \(\pm\)0.062\end{tabular} & \textbf{\begin{tabular}[c]{@{}c@{}}1.000\\ \(\pm\)0.000\end{tabular}} & \begin{tabular}[c]{@{}c@{}}0.983\\ \(\pm\)0.017\end{tabular} & \begin{tabular}[c]{@{}c@{}}0.978\\ \(\pm\)0.019\end{tabular} \\
 & \multirow{-2}{*}{42} & \cellcolor[HTML]{EFEFEF}C4 & \cellcolor[HTML]{EFEFEF}\begin{tabular}[c]{@{}c@{}}0.683\\ \(\pm\)0.186\end{tabular} & \cellcolor[HTML]{EFEFEF}\begin{tabular}[c]{@{}c@{}}0.708\\ \(\pm\)0.058\end{tabular} & \cellcolor[HTML]{EFEFEF}\begin{tabular}[c]{@{}c@{}}0.717\\ \(\pm\)0.057\end{tabular} & \cellcolor[HTML]{EFEFEF}\begin{tabular}[c]{@{}c@{}}0.433\\ \(\pm\)0.198\end{tabular} & \cellcolor[HTML]{EFEFEF}\begin{tabular}[c]{@{}c@{}}0.478\\ \(\pm\)0.063\end{tabular} & \cellcolor[HTML]{EFEFEF}\begin{tabular}[c]{@{}c@{}}0.476\\ \(\pm\)0.063\end{tabular} & \cellcolor[HTML]{EFEFEF}\begin{tabular}[c]{@{}c@{}}0.983\\ \(\pm\)0.034\end{tabular} & \cellcolor[HTML]{EFEFEF}\begin{tabular}[c]{@{}c@{}}0.983\\ \(\pm\)0.016\end{tabular} & \cellcolor[HTML]{EFEFEF}\begin{tabular}[c]{@{}c@{}}0.978\\ \(\pm\)0.018\end{tabular} \\
 & \cellcolor[HTML]{EFEFEF} & C3 & \begin{tabular}[c]{@{}c@{}}0.717\\ \(\pm\)0.180\end{tabular} & \begin{tabular}[c]{@{}c@{}}0.694\\ \(\pm\)0.058\end{tabular} & \begin{tabular}[c]{@{}c@{}}0.693\\ \(\pm\)0.058\end{tabular} & \begin{tabular}[c]{@{}c@{}}0.692\\ \(\pm\)0.185\end{tabular} & \begin{tabular}[c]{@{}c@{}}0.597\\ \(\pm\)0.062\end{tabular} & \begin{tabular}[c]{@{}c@{}}0.644\\ \(\pm\)0.061\end{tabular} & \begin{tabular}[c]{@{}c@{}}0.992\\ \(\pm\)0.022\end{tabular} & \begin{tabular}[c]{@{}c@{}}0.992\\ \(\pm\)0.010\end{tabular} & \begin{tabular}[c]{@{}c@{}}0.973\\ \(\pm\)0.020\end{tabular} \\
 & \multirow{-2}{*}{\cellcolor[HTML]{EFEFEF}48} & \cellcolor[HTML]{EFEFEF}C4 & \cellcolor[HTML]{EFEFEF}\begin{tabular}[c]{@{}c@{}}0.717\\ \(\pm\)0.180\end{tabular} & \cellcolor[HTML]{EFEFEF}\begin{tabular}[c]{@{}c@{}}0.709\\ \(\pm\)0.058\end{tabular} & \cellcolor[HTML]{EFEFEF}\begin{tabular}[c]{@{}c@{}}0.704\\ \(\pm\)0.058\end{tabular} & \cellcolor[HTML]{EFEFEF}\begin{tabular}[c]{@{}c@{}}0.575\\ \(\pm\)0.198\end{tabular} & \cellcolor[HTML]{EFEFEF}\begin{tabular}[c]{@{}c@{}}0.530\\ \(\pm\)0.063\end{tabular} & \cellcolor[HTML]{EFEFEF}\begin{tabular}[c]{@{}c@{}}0.520\\ \(\pm\)0.063\end{tabular} & \cellcolor[HTML]{EFEFEF}\textbf{\begin{tabular}[c]{@{}c@{}}1.000\\ \(\pm\)0.000\end{tabular}} & \cellcolor[HTML]{EFEFEF}\begin{tabular}[c]{@{}c@{}}0.997\\ \(\pm\)0.005\end{tabular} & \cellcolor[HTML]{EFEFEF}\textbf{\begin{tabular}[c]{@{}c@{}}0.996\\ \(\pm\)0.006\end{tabular}} \\ \hline
 &  & C1 & \begin{tabular}[c]{@{}c@{}}0.725\\ \(\pm\)0.179\end{tabular} & \begin{tabular}[c]{@{}c@{}}0.742\\ \(\pm\)0.055\end{tabular} & \begin{tabular}[c]{@{}c@{}}0.736\\ \(\pm\)0.056\end{tabular} & \begin{tabular}[c]{@{}c@{}}0.533\\ \(\pm\)0.200\end{tabular} & \begin{tabular}[c]{@{}c@{}}0.597\\ \(\pm\)0.062\end{tabular} & \begin{tabular}[c]{@{}c@{}}0.569\\ \(\pm\)0.063\end{tabular} & \begin{tabular}[c]{@{}c@{}}0.958\\ \(\pm\)0.061\end{tabular} & \begin{tabular}[c]{@{}c@{}}0.954\\ \(\pm\)0.027\end{tabular} & \begin{tabular}[c]{@{}c@{}}0.951\\ \(\pm\)0.027\end{tabular} \\
 & \multirow{-2}{*}{28} & \cellcolor[HTML]{EFEFEF}C2 & \cellcolor[HTML]{EFEFEF}\begin{tabular}[c]{@{}c@{}}0.783\\ \(\pm\)0.165\end{tabular} & \cellcolor[HTML]{EFEFEF}\begin{tabular}[c]{@{}c@{}}0.735\\ \(\pm\)0.056\end{tabular} & \cellcolor[HTML]{EFEFEF}\begin{tabular}[c]{@{}c@{}}0.788\\ \(\pm\)0.052\end{tabular} & \cellcolor[HTML]{EFEFEF}\begin{tabular}[c]{@{}c@{}}0.867\\ \(\pm\)0.135\end{tabular} & \cellcolor[HTML]{EFEFEF}\begin{tabular}[c]{@{}c@{}}0.832\\ \(\pm\)0.047\end{tabular} & \cellcolor[HTML]{EFEFEF}\begin{tabular}[c]{@{}c@{}}0.865\\ \(\pm\)0.043\end{tabular} & \cellcolor[HTML]{EFEFEF}\begin{tabular}[c]{@{}c@{}}0.917\\ \(\pm\)0.097\end{tabular} & \cellcolor[HTML]{EFEFEF}\begin{tabular}[c]{@{}c@{}}0.863\\ \(\pm\)0.044\end{tabular} & \cellcolor[HTML]{EFEFEF}\begin{tabular}[c]{@{}c@{}}0.886\\ \(\pm\)0.040\end{tabular} \\
 & \cellcolor[HTML]{EFEFEF} & C1 & \begin{tabular}[c]{@{}c@{}}0.7792\\ \(\pm\)0.166\end{tabular} & \begin{tabular}[c]{@{}c@{}}0.767\\ \(\pm\)0.054\end{tabular} & \begin{tabular}[c]{@{}c@{}}0.777\\ \(\pm\)0.053\end{tabular} & \begin{tabular}[c]{@{}c@{}}0.858\\ \(\pm\)0.140\end{tabular} & \begin{tabular}[c]{@{}c@{}}0.863\\ \(\pm\)0.044\end{tabular} & \begin{tabular}[c]{@{}c@{}}0.813\\ \(\pm\)0.049\end{tabular} & \begin{tabular}[c]{@{}c@{}}0.908\\ \(\pm\)0.104\end{tabular} & \begin{tabular}[c]{@{}c@{}}0.913\\ \(\pm\)0.036\end{tabular} & \begin{tabular}[c]{@{}c@{}}0.909\\ \(\pm\)0.036\end{tabular} \\
\multirow{-3}{*}{\begin{tabular}[c]{@{}c@{}}sMRI\_L\\ +\\ sMRI\_R\\ +\\ DTI\_L\\ +\\ DTI\_R\end{tabular}} & \multirow{-2}{*}{\cellcolor[HTML]{EFEFEF}38} & \cellcolor[HTML]{EFEFEF}C2 & \cellcolor[HTML]{EFEFEF}\textbf{\begin{tabular}[c]{@{}c@{}}0.800\\ \(\pm\)0.213\end{tabular}} & \cellcolor[HTML]{EFEFEF}\begin{tabular}[c]{@{}c@{}}0.751\\ \(\pm\)0.077\end{tabular} & \cellcolor[HTML]{EFEFEF}\textbf{\begin{tabular}[c]{@{}c@{}}0.791\\ \(\pm\)0.073\end{tabular}} & \cellcolor[HTML]{EFEFEF}\textbf{\begin{tabular}[c]{@{}c@{}}0.933\\ \(\pm\)0.104\end{tabular}} & \cellcolor[HTML]{EFEFEF}\textbf{\begin{tabular}[c]{@{}c@{}}0.927\\ \(\pm\)0.047\end{tabular}} & \cellcolor[HTML]{EFEFEF}\textbf{\begin{tabular}[c]{@{}c@{}}0.902\\ \(\pm\)0.053\end{tabular}} & \cellcolor[HTML]{EFEFEF}\begin{tabular}[c]{@{}c@{}}0.883\\ \(\pm\)0.149\end{tabular} & \cellcolor[HTML]{EFEFEF}\begin{tabular}[c]{@{}c@{}}0.893\\ \(\pm\)0.055\end{tabular} & \cellcolor[HTML]{EFEFEF}\begin{tabular}[c]{@{}c@{}}0.875\\ \(\pm\)0.059\end{tabular} \\
 &  & C3 & \begin{tabular}[c]{@{}c@{}}0.738\\ \(\pm\)0.176\end{tabular} & \begin{tabular}[c]{@{}c@{}}0.734\\ \(\pm\)0.056\end{tabular} & \begin{tabular}[c]{@{}c@{}}0.763\\ \(\pm\)0.054\end{tabular} & \begin{tabular}[c]{@{}c@{}}0.692\\ \(\pm\)0.185\end{tabular} & \begin{tabular}[c]{@{}c@{}}0.708\\ \(\pm\)0.058\end{tabular} & \begin{tabular}[c]{@{}c@{}}0.723\\ \(\pm\)0.057\end{tabular} & \begin{tabular}[c]{@{}c@{}}0.892\\ \(\pm\)0.116\end{tabular} & \begin{tabular}[c]{@{}c@{}}0.876\\ \(\pm\)0.042\end{tabular} & \begin{tabular}[c]{@{}c@{}}0.903\\ \(\pm\)0.037\end{tabular} \\
 & \multirow{-2}{*}{42} & \cellcolor[HTML]{EFEFEF}C4 & \cellcolor[HTML]{EFEFEF}\begin{tabular}[c]{@{}c@{}}0.721\\ \(\pm\)0.180\end{tabular} & \cellcolor[HTML]{EFEFEF}\begin{tabular}[c]{@{}c@{}}0.743\\ \(\pm\)0.055\end{tabular} & \cellcolor[HTML]{EFEFEF}\begin{tabular}[c]{@{}c@{}}0.744\\ \(\pm\)0.055\end{tabular} & \cellcolor[HTML]{EFEFEF}\begin{tabular}[c]{@{}c@{}}0.650\\ \(\pm\)0.191\end{tabular} & \cellcolor[HTML]{EFEFEF}\begin{tabular}[c]{@{}c@{}}0.699\\ \(\pm\)0.058\end{tabular} & \cellcolor[HTML]{EFEFEF}\begin{tabular}[c]{@{}c@{}}0.674\\ \(\pm\)0.059\end{tabular} & \cellcolor[HTML]{EFEFEF}\begin{tabular}[c]{@{}c@{}}0.900\\ \(\pm\)0.110\end{tabular} & \cellcolor[HTML]{EFEFEF}\begin{tabular}[c]{@{}c@{}}0.909\\ \(\pm\)0.036\end{tabular} & \cellcolor[HTML]{EFEFEF}\begin{tabular}[c]{@{}c@{}}0.914\\ \(\pm\)0.035\end{tabular} \\
 & \cellcolor[HTML]{EFEFEF} & C3 & \begin{tabular}[c]{@{}c@{}}0.788\\ \(\pm\)0.164\end{tabular} & \textbf{\begin{tabular}[c]{@{}c@{}}0.787\\ \(\pm\)0.052\end{tabular}} & \begin{tabular}[c]{@{}c@{}}0.787\\ \(\pm\)0.052\end{tabular} & \begin{tabular}[c]{@{}c@{}}0.792\\ \(\pm\)0.163\end{tabular} & \begin{tabular}[c]{@{}c@{}}0.771\\ \(\pm\)0.053\end{tabular} & \begin{tabular}[c]{@{}c@{}}0.778\\ \(\pm\)0.053\end{tabular} & \textbf{\begin{tabular}[c]{@{}c@{}}0.992\\ \(\pm\)0.022\end{tabular}} & \textbf{\begin{tabular}[c]{@{}c@{}}0.985\\ \(\pm\)0.015\end{tabular}} & \textbf{\begin{tabular}[c]{@{}c@{}}0.986\\ \(\pm\)0.015\end{tabular}} \\
 & \multirow{-2}{*}{\cellcolor[HTML]{EFEFEF}48} & \cellcolor[HTML]{EFEFEF}C4 & \cellcolor[HTML]{EFEFEF}\begin{tabular}[c]{@{}c@{}}0.738\\ \(\pm\)0.176\end{tabular} & \cellcolor[HTML]{EFEFEF}\begin{tabular}[c]{@{}c@{}}0.734\\ \(\pm\)0.056\end{tabular} & \cellcolor[HTML]{EFEFEF}\begin{tabular}[c]{@{}c@{}}0.739\\ \(\pm\)0.056\end{tabular} & \cellcolor[HTML]{EFEFEF}\begin{tabular}[c]{@{}c@{}}0.625\\ \(\pm\)0.194\end{tabular} & \cellcolor[HTML]{EFEFEF}\begin{tabular}[c]{@{}c@{}}0.619\\ \(\pm\)0.061\end{tabular} & \cellcolor[HTML]{EFEFEF}\begin{tabular}[c]{@{}c@{}}0.623\\ \(\pm\)0.061\end{tabular} & \cellcolor[HTML]{EFEFEF}\begin{tabular}[c]{@{}c@{}}0.983\\ \(\pm\)0.034\end{tabular} & \cellcolor[HTML]{EFEFEF}\begin{tabular}[c]{@{}c@{}}0.978\\ \(\pm\)0.019\end{tabular} & \cellcolor[HTML]{EFEFEF}\begin{tabular}[c]{@{}c@{}}0.977\\ \(\pm\)0.019\end{tabular} \\ \hline
\end{tabular}%
}
\caption{AD-MCI classification results. Here \(\alpha \pm \beta\) designation corresponds to the top-mean value of metric and it's \(95\%\) confidence interval. \(\alpha\) - metric value, \([\alpha-\beta, \alpha+\beta]\) - confidence interval.}
\label{table:results_AD_MCI}
\end{table*}
\begin{table*}[]
\centering
\resizebox{\textwidth}{!}{%
\begin{tabular}{c|cc|ccc|ccc|ccc}
\hline
\rowcolor[HTML]{EFEFEF} 
\cellcolor[HTML]{EFEFEF} & \cellcolor[HTML]{EFEFEF} & \cellcolor[HTML]{EFEFEF} & \multicolumn{3}{c|}{\cellcolor[HTML]{EFEFEF}\textbf{\begin{tabular}[c]{@{}c@{}}Top-mean ACC\\ {[}95\% CI{]}\end{tabular}}} & \multicolumn{3}{c|}{\cellcolor[HTML]{EFEFEF}\textbf{\begin{tabular}[c]{@{}c@{}}Top-mean SEN\\ {[}95\% CI{]}\end{tabular}}} & \multicolumn{3}{c}{\cellcolor[HTML]{EFEFEF}\textbf{\begin{tabular}[c]{@{}c@{}}Top-mean SPC\\ {[}95\% CI{]}\end{tabular}}} \\
\rowcolor[HTML]{EFEFEF} 
\multirow{-2}{*}{\cellcolor[HTML]{EFEFEF}\textbf{\begin{tabular}[c]{@{}c@{}}used\\ data\end{tabular}}} & \multirow{-2}{*}{\cellcolor[HTML]{EFEFEF}\textbf{\begin{tabular}[c]{@{}c@{}}ROI\\ size\end{tabular}}} & \multirow{-2}{*}{\cellcolor[HTML]{EFEFEF}\textbf{\begin{tabular}[c]{@{}c@{}}configu-\\ ration\end{tabular}}} & \textbf{test 0} & \textbf{test 1} & \textbf{test 2} & \textbf{test 0} & \textbf{test 1} & \textbf{test 2} & \textbf{test 0} & \textbf{test 1} & \textbf{test 2} \\ \hline
 &  & C1 & \begin{tabular}[c]{@{}c@{}}0.588\\ \(\pm\)0.197\end{tabular} & \begin{tabular}[c]{@{}c@{}}0.558\\ \(\pm\)0.063\end{tabular} & \begin{tabular}[c]{@{}c@{}}0.542\\ \(\pm\)0.063\end{tabular} & \begin{tabular}[c]{@{}c@{}}0.733\\ \(\pm\)0.177\end{tabular} & \begin{tabular}[c]{@{}c@{}}0.710\\ \(\pm\)0.057\end{tabular} & \begin{tabular}[c]{@{}c@{}}0.699\\ \(\pm\)0.058\end{tabular} & \begin{tabular}[c]{@{}c@{}}0.750\\ \(\pm\)0.173\end{tabular} & \begin{tabular}[c]{@{}c@{}}0.725\\ \(\pm\)0.057\end{tabular} & \begin{tabular}[c]{@{}c@{}}0.718\\ \(\pm\)0.057\end{tabular} \\
 & \multirow{-2}{*}{28} & \cellcolor[HTML]{EFEFEF}C2 & \cellcolor[HTML]{EFEFEF}\begin{tabular}[c]{@{}c@{}}0.571\\ \(\pm\)0.198\end{tabular} & \cellcolor[HTML]{EFEFEF}\begin{tabular}[c]{@{}c@{}}0.564\\ \(\pm\)0.063\end{tabular} & \cellcolor[HTML]{EFEFEF}\begin{tabular}[c]{@{}c@{}}0.578\\ \(\pm\)0.063\end{tabular} & \cellcolor[HTML]{EFEFEF}\begin{tabular}[c]{@{}c@{}}\textbf{0.875}\\ \textbf{\(\pm\)0.129}\end{tabular} & \cellcolor[HTML]{EFEFEF}\begin{tabular}[c]{@{}c@{}}\textbf{0.832}\\ \textbf{\(\pm\)0.047}\end{tabular} & \cellcolor[HTML]{EFEFEF}\begin{tabular}[c]{@{}c@{}}\textbf{0.810}\\ \textbf{\(\pm\)0.050}\end{tabular} & \cellcolor[HTML]{EFEFEF}\begin{tabular}[c]{@{}c@{}}0.558\\ \(\pm\)0.199\end{tabular} & \cellcolor[HTML]{EFEFEF}\begin{tabular}[c]{@{}c@{}}0.528\\ \(\pm\)0.063\end{tabular} & \cellcolor[HTML]{EFEFEF}\begin{tabular}[c]{@{}c@{}}0.605\\ \(\pm\)0.062\end{tabular} \\
 & \cellcolor[HTML]{EFEFEF} & C1 & \begin{tabular}[c]{@{}c@{}}0.608\\ \(\pm\)0.195\end{tabular} & \begin{tabular}[c]{@{}c@{}}0.630\\ \(\pm\)0.061\end{tabular} & \begin{tabular}[c]{@{}c@{}}0.603\\ \(\pm\)0.062\end{tabular} & \begin{tabular}[c]{@{}c@{}}0.667\\ \(\pm\)0.189\end{tabular} & \begin{tabular}[c]{@{}c@{}}0.643\\ \(\pm\)0.061\end{tabular} & \begin{tabular}[c]{@{}c@{}}0.667\\ \(\pm\)0.060\end{tabular} & \begin{tabular}[c]{@{}c@{}}0.775\\ \(\pm\)0.167\end{tabular} & \begin{tabular}[c]{@{}c@{}}0.759\\ \(\pm\)0.054\end{tabular} & \begin{tabular}[c]{@{}c@{}}0.797\\ \(\pm\)0.051\end{tabular} \\
 & \multirow{-2}{*}{\cellcolor[HTML]{EFEFEF}38} & \cellcolor[HTML]{EFEFEF}C2 & \cellcolor[HTML]{EFEFEF}\begin{tabular}[c]{@{}c@{}}0.571\\ \(\pm\)0.198\end{tabular} & \cellcolor[HTML]{EFEFEF}\begin{tabular}[c]{@{}c@{}}0.593\\ \(\pm\)0.062\end{tabular} & \cellcolor[HTML]{EFEFEF}\begin{tabular}[c]{@{}c@{}}0.593\\ \(\pm\)0.062\end{tabular} & \cellcolor[HTML]{EFEFEF}\begin{tabular}[c]{@{}c@{}}0.667\\ \(\pm\)0.189\end{tabular} & \cellcolor[HTML]{EFEFEF}\begin{tabular}[c]{@{}c@{}}0.674\\ \(\pm\)0.059\end{tabular} & \cellcolor[HTML]{EFEFEF}\begin{tabular}[c]{@{}c@{}}0.638\\ \(\pm\)0.061\end{tabular} & \cellcolor[HTML]{EFEFEF}\begin{tabular}[c]{@{}c@{}}0.575\\ \(\pm\)0.198\end{tabular} & \cellcolor[HTML]{EFEFEF}\begin{tabular}[c]{@{}c@{}}0.583\\ \(\pm\)0.062\end{tabular} & \cellcolor[HTML]{EFEFEF}\begin{tabular}[c]{@{}c@{}}0.600\\ \(\pm\)0.062\end{tabular} \\
\multirow{-4}{*}{\begin{tabular}[c]{@{}c@{}}sMRI\_L\\ +\\ sMRI\_R\end{tabular}} &  & C3 & \begin{tabular}[c]{@{}c@{}}0.642\\ \(\pm\)0.271\end{tabular} & \begin{tabular}[c]{@{}c@{}}0.614\\ \(\pm\)0.087\end{tabular} & \begin{tabular}[c]{@{}c@{}}0.602\\ \(\pm\)0.088\end{tabular} & \begin{tabular}[c]{@{}c@{}}0.592\\ \(\pm\)0.278\end{tabular} & \begin{tabular}[c]{@{}c@{}}0.608\\ \(\pm\)0.087\end{tabular} & \begin{tabular}[c]{@{}c@{}}0.550\\ \(\pm\)0.089\end{tabular} & \begin{tabular}[c]{@{}c@{}}0.733\\ \(\pm\)0.250\end{tabular} & \begin{tabular}[c]{@{}c@{}}0.673\\ \(\pm\)0.084\end{tabular} & \begin{tabular}[c]{@{}c@{}}0.735\\ \(\pm\)0.079\end{tabular} \\
 & \multirow{-2}{*}{42} & \cellcolor[HTML]{EFEFEF}C4 & \cellcolor[HTML]{EFEFEF}\begin{tabular}[c]{@{}c@{}}0.625\\ \(\pm\)0.194\end{tabular} & \cellcolor[HTML]{EFEFEF}\begin{tabular}[c]{@{}c@{}}0.631\\ \(\pm\)0.061\end{tabular} & \cellcolor[HTML]{EFEFEF}\begin{tabular}[c]{@{}c@{}}0.618\\ \(\pm\)0.062\end{tabular} & \cellcolor[HTML]{EFEFEF}\begin{tabular}[c]{@{}c@{}}0.783\\ \(\pm\)0.165\end{tabular} & \cellcolor[HTML]{EFEFEF}\begin{tabular}[c]{@{}c@{}}0.769\\ \(\pm\)0.053\end{tabular} & \cellcolor[HTML]{EFEFEF}\begin{tabular}[c]{@{}c@{}}0.748\\ \(\pm\)0.055\end{tabular} & \cellcolor[HTML]{EFEFEF}\begin{tabular}[c]{@{}c@{}}0.558\\ \(\pm\)0.199\end{tabular} & \cellcolor[HTML]{EFEFEF}\begin{tabular}[c]{@{}c@{}}0.525\\ \(\pm\)0.063\end{tabular} & \cellcolor[HTML]{EFEFEF}\begin{tabular}[c]{@{}c@{}}0.553\\ \(\pm\)0.063\end{tabular} \\
 & \cellcolor[HTML]{EFEFEF} & C3 & \begin{tabular}[c]{@{}c@{}}\textbf{0.658}\\ \textbf{\(\pm\)0.268}\end{tabular} & \begin{tabular}[c]{@{}c@{}}\textbf{0.657}\\ \textbf{\(\pm\)0.085}\end{tabular} & \begin{tabular}[c]{@{}c@{}}\textbf{0.631}\\ \textbf{\(\pm\)0.086}\end{tabular} & \begin{tabular}[c]{@{}c@{}}0.800\\ \(\pm\)0.213\end{tabular} & \begin{tabular}[c]{@{}c@{}}0.778\\ \(\pm\)0.074\end{tabular} & \begin{tabular}[c]{@{}c@{}}0.713\\ \(\pm\)0.081\end{tabular} & \begin{tabular}[c]{@{}c@{}}\textbf{0.817}\\ \textbf{\(\pm\)0.201}\end{tabular} & \begin{tabular}[c]{@{}c@{}}\textbf{0.788}\\ \textbf{\(\pm\)0.073}\end{tabular} & \begin{tabular}[c]{@{}c@{}}\textbf{0.788}\\ \textbf{\(\pm\)0.073}\end{tabular} \\
 & \multirow{-2}{*}{\cellcolor[HTML]{EFEFEF}48} & \cellcolor[HTML]{EFEFEF}C4 & \cellcolor[HTML]{EFEFEF}\begin{tabular}[c]{@{}c@{}}0.542\\ \(\pm\)0.199\end{tabular} & \cellcolor[HTML]{EFEFEF}\begin{tabular}[c]{@{}c@{}}0.502\\ \(\pm\)0.063\end{tabular} & \cellcolor[HTML]{EFEFEF}\begin{tabular}[c]{@{}c@{}}0.518\\ \(\pm\)0.063\end{tabular} & \cellcolor[HTML]{EFEFEF}\begin{tabular}[c]{@{}c@{}}0.683\\ \(\pm\)0.186\end{tabular} & \cellcolor[HTML]{EFEFEF}\begin{tabular}[c]{@{}c@{}}0.696\\ \(\pm\)0.058\end{tabular} & \cellcolor[HTML]{EFEFEF}\begin{tabular}[c]{@{}c@{}}0.671\\ \(\pm\)0.060\end{tabular} & \cellcolor[HTML]{EFEFEF}\begin{tabular}[c]{@{}c@{}}0.600\\ \(\pm\)0.196\end{tabular} & \cellcolor[HTML]{EFEFEF}\begin{tabular}[c]{@{}c@{}}0.560\\ \(\pm\)0.063\end{tabular} & \cellcolor[HTML]{EFEFEF}\begin{tabular}[c]{@{}c@{}}0.618\\ \(\pm\)0.062\end{tabular} \\ \hline
 &  & C1 & \begin{tabular}[c]{@{}c@{}}0.538\\ \(\pm\)0.200\end{tabular} & \begin{tabular}[c]{@{}c@{}}0.532\\ \(\pm\)0.063\end{tabular} & \begin{tabular}[c]{@{}c@{}}0.517\\ \(\pm\)0.063\end{tabular} & \begin{tabular}[c]{@{}c@{}}0.767\\ \(\pm\)0.169\end{tabular} & \begin{tabular}[c]{@{}c@{}}0.788\\ \(\pm\)0.052\end{tabular} & \begin{tabular}[c]{@{}c@{}}0.693\\ \(\pm\)0.058\end{tabular} & \begin{tabular}[c]{@{}c@{}}0.500\\ \(\pm\)0.200\end{tabular} & \begin{tabular}[c]{@{}c@{}}0.518\\ \(\pm\)0.063\end{tabular} & \begin{tabular}[c]{@{}c@{}}0.542\\ \(\pm\)0.063\end{tabular} \\
 & \multirow{-2}{*}{28} & \cellcolor[HTML]{EFEFEF}C2 & \cellcolor[HTML]{EFEFEF}\begin{tabular}[c]{@{}c@{}}0.538\\ \(\pm\)0.200\end{tabular} & \cellcolor[HTML]{EFEFEF}\begin{tabular}[c]{@{}c@{}}0.563\\ \(\pm\)0.063\end{tabular} & \cellcolor[HTML]{EFEFEF}\begin{tabular}[c]{@{}c@{}}0.548\\ \(\pm\)0.063\end{tabular} & \cellcolor[HTML]{EFEFEF}\begin{tabular}[c]{@{}c@{}}0.800\\ \(\pm\)0.160\end{tabular} & \cellcolor[HTML]{EFEFEF}\begin{tabular}[c]{@{}c@{}}0.777\\ \(\pm\)0.053\end{tabular} & \cellcolor[HTML]{EFEFEF}\begin{tabular}[c]{@{}c@{}}0.713\\ \(\pm\)0.057\end{tabular} & \cellcolor[HTML]{EFEFEF}\begin{tabular}[c]{@{}c@{}}0.500\\ \(\pm\)0.200\end{tabular} & \cellcolor[HTML]{EFEFEF}\begin{tabular}[c]{@{}c@{}}0.472\\ \(\pm\)0.063\end{tabular} & \cellcolor[HTML]{EFEFEF}\begin{tabular}[c]{@{}c@{}}0.504\\ \(\pm\)0.063\end{tabular} \\
 & \cellcolor[HTML]{EFEFEF} & C1 & \begin{tabular}[c]{@{}c@{}}0.542\\ \(\pm\)0.199\end{tabular} & \begin{tabular}[c]{@{}c@{}}0.547\\ \(\pm\)0.063\end{tabular} & \begin{tabular}[c]{@{}c@{}}0.556\\ \(\pm\)0.063\end{tabular} & \begin{tabular}[c]{@{}c@{}}0.758\\ \(\pm\)0.171\end{tabular} & \begin{tabular}[c]{@{}c@{}}0.738\\ \(\pm\)0.056\end{tabular} & \begin{tabular}[c]{@{}c@{}}0.706\\ \(\pm\)0.058\end{tabular} & \begin{tabular}[c]{@{}c@{}}0.417\\ \(\pm\)0.197\end{tabular} & \begin{tabular}[c]{@{}c@{}}0.403\\ \(\pm\)0.062\end{tabular} & \begin{tabular}[c]{@{}c@{}}0.501\\ \(\pm\)0.063\end{tabular} \\
\multirow{-3}{*}{\begin{tabular}[c]{@{}c@{}}sMRI\_L\\ +\\ sMRI\_R\\ +\\ DTI\_L\\ +\\ DTI\_R\end{tabular}} & \multirow{-2}{*}{\cellcolor[HTML]{EFEFEF}38} & \cellcolor[HTML]{EFEFEF}C2 & \cellcolor[HTML]{EFEFEF}\begin{tabular}[c]{@{}c@{}}\textbf{0.625}\\ \textbf{\(\pm\)0.194}\end{tabular} & \cellcolor[HTML]{EFEFEF}\begin{tabular}[c]{@{}c@{}}\textbf{0.613}\\ \textbf{\(\pm\)0.062}\end{tabular} & \cellcolor[HTML]{EFEFEF}\begin{tabular}[c]{@{}c@{}}\textbf{0.612}\\ \textbf{\(\pm\)0.062}\end{tabular} & \cellcolor[HTML]{EFEFEF}\begin{tabular}[c]{@{}c@{}}\textbf{0.858}\\ \textbf{\(\pm\)0.140}\end{tabular} & \cellcolor[HTML]{EFEFEF}\begin{tabular}[c]{@{}c@{}}\textbf{0.834}\\ \textbf{\(\pm\)0.047}\end{tabular} & \cellcolor[HTML]{EFEFEF}\begin{tabular}[c]{@{}c@{}}\textbf{0.792}\\ \textbf{\(\pm\)0.051}\end{tabular} & \cellcolor[HTML]{EFEFEF}\begin{tabular}[c]{@{}c@{}}0.458\\ \(\pm\)0.199\end{tabular} & \cellcolor[HTML]{EFEFEF}\begin{tabular}[c]{@{}c@{}}0.441\\ \(\pm\)0.063\end{tabular} & \cellcolor[HTML]{EFEFEF}\begin{tabular}[c]{@{}c@{}}0.520\\ \(\pm\)0.063\end{tabular} \\
 &  & C3 & \begin{tabular}[c]{@{}c@{}}0.621\\ \(\pm\)0.194\end{tabular} & \begin{tabular}[c]{@{}c@{}}0.628\\ \(\pm\)0.061\end{tabular} & \begin{tabular}[c]{@{}c@{}}0.608\\ \(\pm\)0.062\end{tabular} & \begin{tabular}[c]{@{}c@{}}0.708\\ \(\pm\)0.182\end{tabular} & \begin{tabular}[c]{@{}c@{}}0.759\\ \(\pm\)0.054\end{tabular} & \begin{tabular}[c]{@{}c@{}}0.690\\ \(\pm\)0.059\end{tabular} & \begin{tabular}[c]{@{}c@{}}\textbf{0.600}\\ \textbf{\(\pm\)0.196}\end{tabular} & \begin{tabular}[c]{@{}c@{}}\textbf{0.570}\\ \textbf{\(\pm\)0.063}\end{tabular} & \begin{tabular}[c]{@{}c@{}}\textbf{0.624}\\ \textbf{\(\pm\)0.061}\end{tabular} \\
 & \multirow{-2}{*}{42} & \cellcolor[HTML]{EFEFEF}C4 & \cellcolor[HTML]{EFEFEF}\begin{tabular}[c]{@{}c@{}}0.513\\ \(\pm\)0.200\end{tabular} & \cellcolor[HTML]{EFEFEF}\begin{tabular}[c]{@{}c@{}}0.537\\ \(\pm\)0.063\end{tabular} & \cellcolor[HTML]{EFEFEF}\begin{tabular}[c]{@{}c@{}}0.531\\ \(\pm\)0.063\end{tabular} & \cellcolor[HTML]{EFEFEF}\begin{tabular}[c]{@{}c@{}}0.692\\ \(\pm\)0.185\end{tabular} & \cellcolor[HTML]{EFEFEF}\begin{tabular}[c]{@{}c@{}}0.728\\ \(\pm\)0.056\end{tabular} & \cellcolor[HTML]{EFEFEF}\begin{tabular}[c]{@{}c@{}}0.673\\ \(\pm\)0.060\end{tabular} & \cellcolor[HTML]{EFEFEF}\begin{tabular}[c]{@{}c@{}}0.408\\ \(\pm\)0.197\end{tabular} & \cellcolor[HTML]{EFEFEF}\begin{tabular}[c]{@{}c@{}}0.424\\ \(\pm\)0.063\end{tabular} & \cellcolor[HTML]{EFEFEF}\begin{tabular}[c]{@{}c@{}}0.453\\ \(\pm\)0.063\end{tabular} \\
 & \cellcolor[HTML]{EFEFEF} & C3 & \begin{tabular}[c]{@{}c@{}}0.588\\ \(\pm\)0.197\end{tabular} & \begin{tabular}[c]{@{}c@{}}0.585\\ \(\pm\)0.062\end{tabular} & \begin{tabular}[c]{@{}c@{}}0.573\\ \(\pm\)0.063\end{tabular} & \begin{tabular}[c]{@{}c@{}}0.708\\ \(\pm\)0.182\end{tabular} & \begin{tabular}[c]{@{}c@{}}0.697\\ \(\pm\)0.058\end{tabular} & \begin{tabular}[c]{@{}c@{}}0.638\\ \(\pm\)0.061\end{tabular} & \begin{tabular}[c]{@{}c@{}}0.567\\ \(\pm\)0.198\end{tabular} & \begin{tabular}[c]{@{}c@{}}0.520\\ \(\pm\)0.063\end{tabular} & \begin{tabular}[c]{@{}c@{}}0.604\\ \(\pm\)0.062\end{tabular} \\
 & \multirow{-2}{*}{\cellcolor[HTML]{EFEFEF}48} & \cellcolor[HTML]{EFEFEF}C4 & \cellcolor[HTML]{EFEFEF}\begin{tabular}[c]{@{}c@{}}0.554\\ \(\pm\)0.199\end{tabular} & \cellcolor[HTML]{EFEFEF}\begin{tabular}[c]{@{}c@{}}0.543\\ \(\pm\)0.063\end{tabular} & \cellcolor[HTML]{EFEFEF}\begin{tabular}[c]{@{}c@{}}0.552\\ \(\pm\)0.063\end{tabular} & \cellcolor[HTML]{EFEFEF}\begin{tabular}[c]{@{}c@{}}0.725\\ \(\pm\)0.179\end{tabular} & \cellcolor[HTML]{EFEFEF}\begin{tabular}[c]{@{}c@{}}0.719\\ \(\pm\)0.057\end{tabular} & \cellcolor[HTML]{EFEFEF}\begin{tabular}[c]{@{}c@{}}0.707\\ \(\pm\)0.058\end{tabular} & \cellcolor[HTML]{EFEFEF}\begin{tabular}[c]{@{}c@{}}0.467\\ \(\pm\)0.200\end{tabular} & \cellcolor[HTML]{EFEFEF}\begin{tabular}[c]{@{}c@{}}0.450\\ \(\pm\)0.063\end{tabular} & \cellcolor[HTML]{EFEFEF}\begin{tabular}[c]{@{}c@{}}0.471\\ \(\pm\)0.063\end{tabular} \\ \hline
\end{tabular}%
}
\caption{MCI-NC classification results. Here \(\alpha \pm \beta\) designation corresponds to the top-mean value of metric and it's \(95\%\) confidence interval. \(\alpha\) - metric value, \([\alpha-\beta, \alpha+\beta]\) - confidence interval.}
\label{table:results_MCI_NC}
\end{table*}
In this work we analyze the dependency of used data and network configuration on the efficiency of Alzheimer disease detection. We run a number of experiments with different configurations: used image modalities (sMRI, DTI), ROI sizes, number of convolutional layers, number of convolutions in each layer and compare them. As in previous joint works \cite{Aderghal} we train and evaluate 3 binary classifiers: AD-NC, AD-MCI, MCI-NC. The obtained results are shown in tables \ref{table:results_AD_NC}, \ref{table:results_AD_MCI}, \ref{table:results_MCI_NC}.

\begin{figure}[h]
\centering\includegraphics[width=\linewidth]{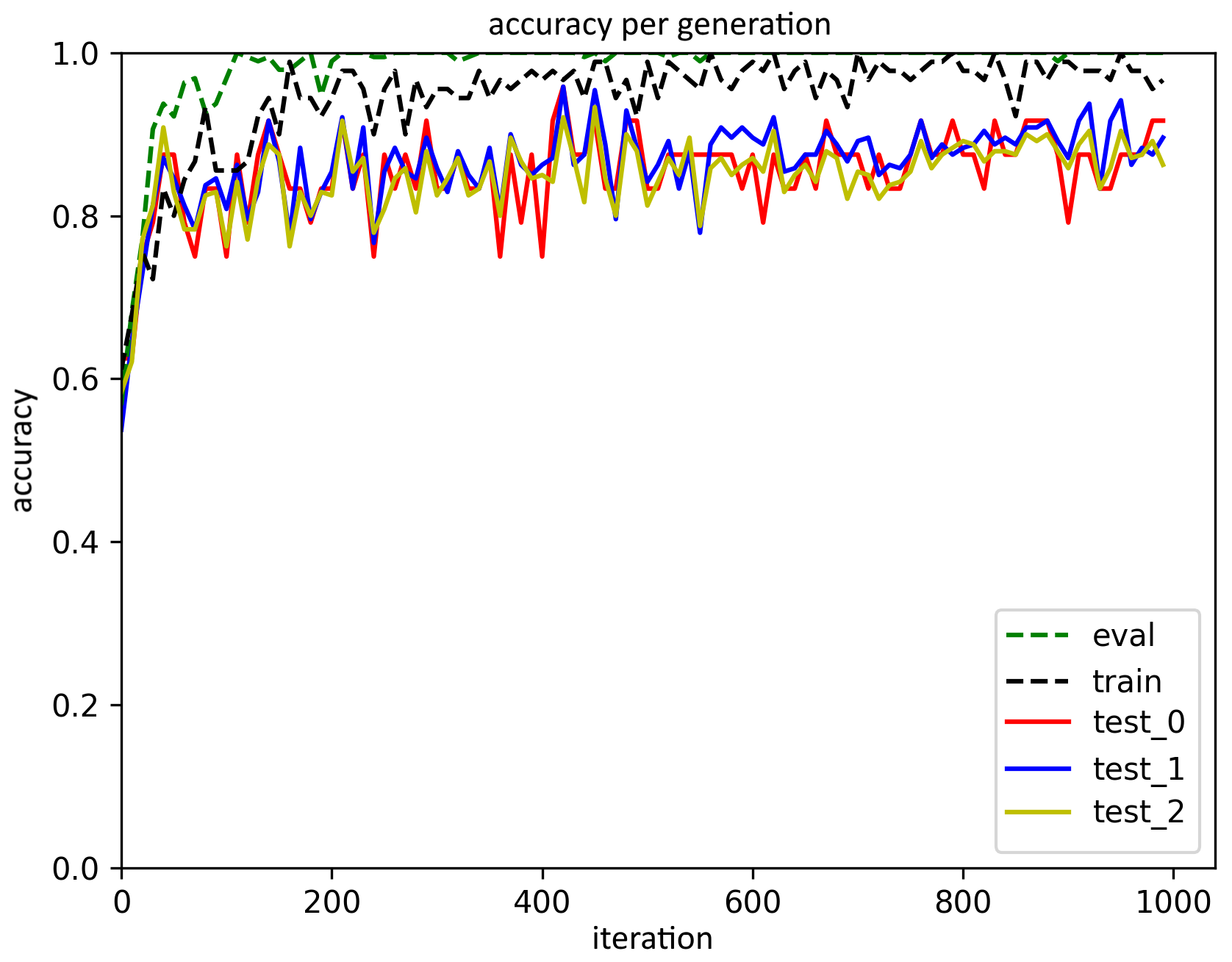}
\caption{An example of accuracy plot for 1000 iteration during training the network for train, validation, test 0, test 1 and test 2 sets.}
\label{fig:accuracy}
\end{figure}

To evaluate and score each experiment we use accuracy as a reference metric. As the database we use in this paper is not large enough, the typical accuracy curve for the test set among iterations is not smooth (Fig.\ref{fig:accuracy}). To eliminate this problem and densely characterize each experiment we propose to calculate mean accuracy for every \(s\) sequential iterations and find the maximum. Let's call this value a top-mean accuracy. In this work an interval of \(s=100\) iterations is chosen. For a more precise description of accuracy curve we also calculate accuracy variance on the same interval with the top-mean value.

Along with accuracy (ACC) we also report the values of sensitivity (SEN) and specificity (SPC). All described metrics are calculated using a top-mean approach. We should also notice the absence of commonly used balanced accuracy (BAC) metric. That is because we use an already balanced test set, as the number of patients used for testing in each class is the same (Table \ref{table:samples}). So in the current case all reported accuracy values in this work are are equal to the balanced accuracy values.

To perform an interval estimation of the classification we report a 95\% confidence interval for all described metrics using Wilson score interval \cite{wilson_1,wilson_2}. The confidence interval is calculated as:
\[
	CI = val \pm \theta \cdot \sqrt[]{\frac{val (1-val)}{n}},
\]
where \(\theta\) is a constant corresponding to confidence range (in case of 95\% range \(\theta=1.96\)), \(n\) is a number of samples in the set and \(val\) is a value of metric for which we calculate a confidence interval.
Here we should also notice, that the width of the confidence interval depends on the number of samples in the set: the more samples there is in the set, the shorter confidence interval would be. There by in our case we get shorter confidence intervals for test 1 and test 2 sets than for the test 0 set because of augmentation.

As it was discussed earlier for each ROI and each modality we perform a base pipeline of convolutions and then do a late fusion. To  compare sMRI and DTI modalities and analyze their applicability to Alzheimer disease detection we consider neural networks with the following pipeline inputs (the corresponding abbreviations used below are given at the beginning of the lines):
\begin{enumerate}
\item \textit{DTI\_L+DTI\_R}: left hippocampus on MD-DTI and right hippocampus on MD-DTI images
\item \textit{sMRI\_L+sMRI\_R}: left hippocampus on sMRI and right hippocampus on sMRI images
\item \textit{sMRI\_L+sMRI\_R+DTI\_L+DTI\_R}: left hippocampus on sMRI, right hippocampus on sMRI, left hippocampus on MD-DTI and right hippocampus on MD-DTI images
\item \textit{sMRI\_LR+DTI\_LR}: left-right hippocampus on sMRI and left-right hippocampus on MD-DTI images
\end{enumerate}
Here left-right hippocampus corresponds to the data set obtained by uniting left hippocampus ROI and mirror-flipped right hippocampus ROI. This is justified from medical point of view as left and right lobes of hippocampus represent symmetrical structures. An example of used network is shown in Figure \ref{fig:NN_architecture}.

During the experiments it was found that AD-NC and AD-MCI binary classifiers achieve the best classification scores with the third type of input (left hippocampus on MD-DTI + right hippocampus on MD-DTI + left hippocampus on sMRI + right hippocampus on sMRI). Slightly inferior results are obtained with data fusion in the most difficult MCI-NC case. It can be explained by the fact that adding a less informative DTI modality in conditions of small amount of data leads to making the weights of fully-connected layers of the network more noisy and thereby worsening the final result.

The results obtained using the first type of input on a single DTI modality (left hippocampus on MD-DTI and right hippocampus on MD-DTI) were the worst for all three classifiers, so we do not add them into tables \ref{table:results_AD_NC}, \ref{table:results_AD_MCI}, \ref{table:results_MCI_NC}. Moreover the results obtained with the fourth type of input (left-right hippocampus on sMRI and left-right hippocampus on MD-DTI) were in almost cases better than in case of using sMRI data only (second type of input), but worser than using data fusion for each ROI separately (third type of input). Presumably this happens due to the fact that although left and right hippocampal structures are symmetrical they are not identical. So treating them separately gives a greater effect than using them simultaneously. For the described reason we do not include this type of input in the results tables \ref{table:results_AD_NC}, \ref{table:results_AD_MCI}, \ref{table:results_MCI_NC}.

The results also demonstrate that the size of ROIs does matter. The usage of bigger ROIs (\(38 \times 38 \times 38\) and \(48 \times 48 \times 48\) voxels) in combination with a deeper network architecture leads to better classification results. So for example we achieved a classification accuracy of 0.967 using \(48 \times 48 \times 48\) ROI and a 6-layered network with data fusion. A combination of \(38 \times 38 \times 38\) ROIs with 5-layered \(C2\) architecture also demonstrated a good level of performance in all three classification cases.

We should also discuss the impact of test set augmentation on the classification results. As it can be seen from Fig. \ref{fig:accuracy} and tables \ref{table:results_AD_NC}, \ref{table:results_AD_MCI}, \ref{table:results_MCI_NC}, the usage of more augmented set leads to a smoother accuracy curve, but also slightly worsens the classification results in all cases.

All in all we succeeded to achieve the classification accuracy of 0.967, 0.8 and 0.658 for AD-NC, AD-MCI and MCI-NC classification problems respectively.

The proposed method was implemented using the TensorFlow \cite{tensorflow} framework.
The experiments were performed on two configurations: Intel Core i7-6700HQ with Nvidia GeForce GTX 960M and Intel Core i7-7700HQ CPU with Nvidia GeForce GTX 1070 GPU.

\section{Discussion and Conclusion}
\label{S:Conclusion}
As it can be seen from the Table \ref{comparison_table} relatively new feature-based and neural network-based methods demonstrate very good level of performance compared to the classical volumetric methods that are performed manually by medical experts.

It should be mentioned, that the direct comparison of the reviewed algorithms for Alzheimer's disease diagnostics is impossible. The proposed results were obtained using images from several databases and in different quantities (see Table \ref{comparison_table}). Moreover different classification problems were challenged: although most papers focus on the 3-class AD/MCI/NC  binary classification, some of them consider only 2-class AD/NC classification \cite{c4_sparse_encoder_Korea, c7.1_DeepAd_Canada, c7.2_DeeapAd_Canada, c7.3_DeeapAd_Canada} and even 4-class AD/eMCI/lMCI/NC classification \cite{c3_Skolkovo}. Also \cite{c5_demnet_Philippines,c6_deep_ensemble_sparse_regr_Korea} deserve special attention as the authors try to solve a  problem in demand of prediction of Alzheimer converters. Nevertheless, we can state that our results confirm the general trend: with Deep Neural Networks on 3D volumes and fusing different modalities, we achieve scores of accuracy higher then 0.9. This makes us think that the CNN - based classification can indeed be used for real-world CAD systems in large cohort screening. In this paper we focused on only one biomarler ROI, the hippocampal ROI. Nevertheless, accordingly to previous research \cite{f3_Jenny_pcc} it is interesting to add other ROIs known to be deteriorated due to AD.

\section{Acknowledgements}
Data  collection  and  sharing  for  this  project  was  funded  by  the  Alzheimer's  Disease Neuroimaging  Initiative  (ADNI)  (National  Institutes  of  Health  Grant  U01  AG024904)  and DOD  ADNI  (Department  of  Defense  award  number  W81XWH-12-2-0012). ADNI  is  funded by  the  National Institute  on  Aging,  the  National  Institute of  Biomedical  Imaging  and Bioengineering, and through generous contributions from the following: AbbVie, Alzheimer’s Association;  Alzheimer’s  Drug  Discovery  Foundation; Araclon Biotech; BioClinica, Inc.; Biogen; Bristol-Myers Squibb Company; CereSpir, Inc.; Cogstate;Eisai Inc.; Elan Pharmaceuticals, Inc.; Eli Lilly and Company; EuroImmun;  F.Hoffmann-La  Roche  Ltd  and its  affiliated  company  Genentech, Inc.;  Fujirebio;  GE Healthcare;  IXICO  Ltd.;  Janssen Alzheimer    Immunotherapy  Research  and   Development, LLC.; Johnson   and  Johnson Pharmaceutical  Research  and Development LLC.; Lumosity; Lundbeck; Merck and Co., Inc.; Meso Scale Diagnostics, LLC.; NeuroRx  Research; Neurotrack Technologies; Novartis Pharmaceuticals Corporation; Pfizer Inc.; Piramal Imaging; Servier; Takeda Pharmaceutical Company; and Transition Therapeutics.
The  Canadian  Institutes  of  Health  Research  is providing  funds  to  support  ADNI  clinical  sites  in  Canada.  Private  sector  contributions  are facilitated by the Foundation for the National Institutes of Health (\url{www.fnih.org}). The grantee organization is the Northern California Institute for Research and Education, and the study is coordinated by the Alzheimer’s Therapeutic Research Institute at the University of Southern California.  ADNI  data  are  disseminated  by  the  Laboratory  for  Neuroimaging  at  the University of Southern California.

This research was supported by Ostrogradsky scholarship grant 2017 established by French Embassy in Russia and TOUBKAL French-Morocco research grant Alclass. We thank Dr. Pierrick Coup\'e from LABRI UMR 5800 University of Bordeaux/CNRS/Bordeaux-IPN who provided insight and expertise that greatly assisted the research.

\bibliographystyle{unsrt}
\bibliography{biblio}

\end{document}